\documentclass[a4paper,fleqn,]{cas-dc}
\usepackage{comment} 
\usepackage{caption}
\usepackage[numbers]{natbib}
\usepackage{ulem} 
\usepackage{latexsym}
\usepackage{multirow}
\usepackage{yhmath}
\usepackage{bm}
\usepackage{graphicx}
\usepackage{latexsym}

\useunder{\uline}{\ul}{}
\usepackage{url}
\usepackage{pifont}
\usepackage{makecell}
\usepackage{lscape}
\usepackage{diagbox}
\usepackage[ruled,linesnumbered]{algorithm2e}
\usepackage{amsmath,amssymb,amsfonts,dsfont}
\usepackage{subfigure}
\usepackage{algorithmic}
\usepackage{hyperref}
\hypersetup{
	colorlinks=true,
	linkcolor=black,
	filecolor=cyan,
	urlcolor=black,
	citecolor=black,
}
\usepackage{listings}
\usepackage{xcolor}
\usepackage{xpatch}
\long\def\stmdocstextcolor#1#2{#2}
\def\stmdocscolor#1{}

\def\tsc#1{\csdef{#1}{\textsc{\lowercase{#1}}\xspace}}
\tsc{WGM}
\tsc{QE}
\tsc{EP}
\tsc{PMS}
\tsc{BEC}
\tsc{DE}

\begin{document}
\let\WriteBookmarks\relax
\def\floatpagepagefraction{0.8}
\def\textpagefraction{.001}

\shorttitle{ReZG}

\shortauthors{Jiang et~al.}
\title [mode = title]{ReZG: Retrieval-Augmented Zero-Shot Counter Narrative Generation for Hate Speech}    

\author[1]{Shuyu Jiang}[style=chinese]

\author[1,3]{Wenyi Tang}[style=chinese]

\author[1,2,3]{Xingshu Chen}[style=chinese]

\author[1,3]{Rui Tang}[style=chinese]
\ead{tangrscu@scu.edu.cn}
\cormark[1]

\author[1,3]{Haizhou Wang}[style=chinese]
\author[2,3]{Wenxian Wang}[style=chinese]

\address[1]{School of Cyber Science and Engineering, Sichuan University,Chengdu,China}
\address[2]{Cyber Science Research Institute, Sichuan University, Chengdu, China}

\address[3]{Key Laboratory of Data Protection and Intelligent Management (Sichuan University), Ministry of Education, Chengdu, China}

\cortext[cor1]{Corresponding author}

\begin{abstract}
The proliferation of {h}ate {s}peech (HS) on social media poses a serious threat to societal security. Automatic {c}ounter {n}arrative (CN) generation, as an active strategy for HS intervention, has garnered increasing attention in recent years. Existing methods for automatically generating CNs mainly rely on re-training or fine-tuning pre-trained language models (PLMs) on human-curated CN corpora. Unfortunately, the annotation speed of CN corpora cannot keep up with the growth of HS targets, while generating specific and effective CNs for unseen targets remains a significant challenge for the model. To tackle this issue, we propose \textbf{Re}trieval-Augmented \textbf{Z}ero-shot \textbf{G}eneration (ReZG) to generate CNs with high-specificity for unseen targets. Specifically, we propose a multi-dimensional hierarchical retrieval method that integrates stance, semantics, and fitness, extending the retrieval metric from single dimension to multiple dimensions suitable for the knowledge that refutes HS.
Then, we implement an energy-based constrained decoding mechanism that enables PLMs to use differentiable knowledge preservation, countering, and fluency constraint functions instead of in-target CNs as control signals for generation, thereby achieving zero-shot CN generation. With the above techniques, ReZG can integrate external knowledge flexibly and improve the specificity of CNs.
Experimental results show that ReZG exhibits stronger generalization capabilities and outperforms strong baselines with significant improvements of {\bf 2.0\%+} in the relevance and {\bf 4.5\%+} in the countering success rate metrics.

\textbf{Warning}:\textit{ Examples in this paper may contain offensive or upsetting content.}
\end{abstract}

\maketitle

\section{Introduction}
The prevalence of social media platforms and the power of  anonymity
\stmdocstextcolor{black}{have} triggered an upsurge and proliferation of online
\stmdocstextcolor{black}{h}ate \stmdocstextcolor{black}{s}peech
(HS)~\citep{JAHAN2023126232}. This phenomenon is seriously endangering the
security of the entire society~\citep{williams2019hatred}, since online HS not
only \stmdocstextcolor{black}{severely} affects the mental health of victims,
but \stmdocstextcolor{black}{may also progress} into offline street hate crimes
\stmdocstextcolor{black}{such as} the Pittsburg shooting, the anti-Muslim mob
violence in Sri Lanka and violence at the US Capitol
~\citep{mathew2020hate,SENGUPTA2022598}.

The \stmdocstextcolor{black}{c}ounter \stmdocstextcolor{black}{n}arrative (CN), as a promising countermeasure against online HS without restricting freedom of speech, is being advocated and used by many \stmdocstextcolor{black}{n}on-\stmdocstextcolor{black}{g}overnmental \stmdocstextcolor{black}{o}rganizations (NGOs)\footnote{\url{https://getthetrollsout.org/the-project}} \footnote{\url{https://dangerousspeech.org/}}.  

A CN is a non-aggressive response that utilizes logic, facts, or alternative
perspectives to refute
HS~\citep{schieb2016governing,fanton2021human,tekiroglu2022using}, aiming to
de-escalate hateful sentiment and foster mutual understanding by exchanging
opinions. 
 Initial CNs are mainly acquired through human curation. As the proliferation
of online HS surpasses the capacity for manual CN creation, automated CN
generation \stmdocstextcolor{black}{is becoming} increasingly vital and has
attracted growing attention in the field of \stmdocstextcolor{black}{n}atural
\stmdocstextcolor{black}{l}anguage \stmdocstextcolor{black}{g}eneration
(NLG)~\citep{tekiroglu2022using,radford2019language,chung2020italian,ashida2022towards,raffel2020exploring,gupta2023counterspeeches}. 
Typically, \stmdocstextcolor{black}{automated CN generation} is mainly achieved
through retraining or fine-tuning \stmdocstextcolor{black}{pre-trained language
models} (PLMs) on human-curated CN
corpora~\citep{tekirouglu2020generating,tekiroglu2022using,chung2021towards}. 

\stmdocstextcolor{black}{In general}, supporting materials for effective
refutation vary depending on the targets of HS, which are arising
\stmdocstextcolor{black}{continuously} with the rapid evolution of topics on
online social media~\citep{ludwig-etal-2022-improving}.
Facing this ever-increasing number of HS targets, annotated CN corpora for new
targets remain deficient. When the model is unexposed to the CNs of new targets
during the training process, the specificity and effectiveness of the generated
CNs will decrease \stmdocstextcolor{black}{significantly} within
\stmdocstextcolor{black}{these} unseen
targets~\citep{tekiroglu2022using,bonaldi2023weigh} .

As depicted in 
Table \tabref{tab_exp:in-out-example}, it illustrates in-target and out-target CNs against the same HS, generated by models in the above two different conditions.
These two models are identical in all aspects except for the training datasets. The training
dataset of CN1 model includes CNs with the target of ``women'', whereas the training dataset of CN2 model excludes these.
Evidently, CN1 can directly refute the core points of ``\texttt{women are childlike}'' within the HS, demonstrating a stronger specificity and countering capability. \stmdocstextcolor{black}{Thus, it} is an effective CN. Such knowledge that can be used to refute HS is called counter-knowledge, which can be acquired from statistics, facts, examples, or alternative perspectives. 
\stmdocstextcolor{black}{In contrast}, CN2 deviates from the core content of the HS, failing to provide pertinent countering logic, facts, or examples. Its lack of specificity renders it nearly an invalid CN.

\begin{table}[htb]
\centering
\caption{An illustrative example of CNs generated by the models identical in all aspects except for the training dataset. The training dataset of CN1 model encompasses CNs with the target of ``\texttt{women}'', whereas the training dataset of CN2 model excludes these ones.}
 \begin{tabular}{l|p{5cm}}
    \hline                                        
    \textbf{Hate Speech} &Women are basically childlike, they remain this way most of their lives.   They are a moral failing for our society. \\
    \hline
     \textbf{CN1 (In-target)} &So you think women are childlike because they are submissive?  No they are not.  They are strong and independent, just like men. \\
     \hline
  \textbf{CN2 (Out-target)} & I think it would be more appropriate to call a person by their name instead of calling them by their gender.\\
    \hline
  \end{tabular}
\label{tab_exp:in-out-example}
\end{table}

To address this issue, this \stmdocstextcolor{black}{study} proposes a \textbf{Re}trieval-augmented \textbf{Z}ero-shot \textbf{G}eneration method for CNs (ReZG), which can generate CNs with high-specificity \stmdocstextcolor{black}{by} retrieving and mapping counter-knowledge from external supporting materials into CNs. We utilize \stmdocstextcolor{black}{a} PLM that has not been trained on the CN generation task as the generator, coupled with an energy-based constrained decoding mechanism to achieve control over the generated sequence.

\begin{figure*}[htbp!]
  \centering
  \includegraphics[width=0.9\textwidth]{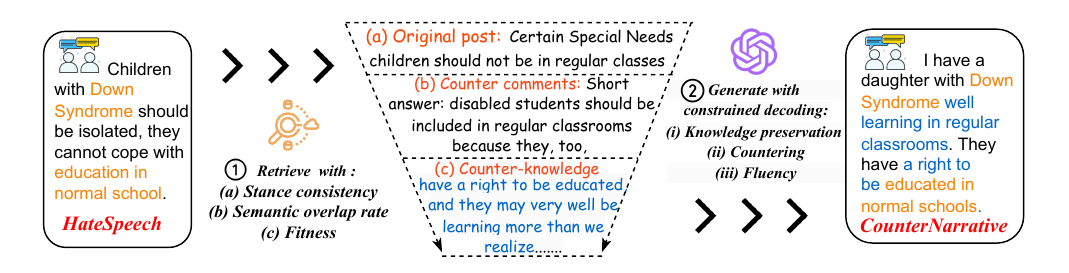}
  \caption{ReZG method workflow. Blue words represent the retrieved counter-knowledge, and yellow words represent the refuted points of HS.}\label{pic:intro}
\end{figure*}

As shown in Fig.\ref{pic:intro}, ReZG comprises retrieval and generation \stmdocstextcolor{black}{modules}. The retrieval module makes up for the lack of unseen-target-related information by retrieving relevant counter-knowledge. The generation module uses the knowledge preservation, countering and fluency constraints instead of \stmdocstextcolor{black}{in-target} CNs as the control signal to generate a specific, cogent and fluent CN, \stmdocstextcolor{black}{thereby} achieving the zero-shot generation.

Generally speaking, traditional information retrieval methods  mainly rely on semantic similarity to obtain relevant information, which is insufficient for acquiring counter-knowledge. CNs should hold the opposite views to HS, involving aspects like stance in addition to semantics. Therefore, this \stmdocstextcolor{black}{study} customizes a multi-dimensional hierarchical retrieval method to gradually screen counter-knowledge from coarse-grained to fine-grained from the perspectives of stance consistency, semantic overlap rate and fitness to HS, to reduce \stmdocstextcolor{black}{the} non-countering information in traditional semantic-related retrieval results.

In the generation module, the generation process becomes uncontrollable since it is \stmdocstextcolor{black}{difficult} for the model to learn the complex mapping relationships between counter-knowledge and target CNs without corresponding annotated datasets. To fill this gap, we implement a multi-constraint decoding mechanism based on energy function, through introducing three differentiable constraint functions to the decoding stage, to facilitate these mappings while improving the specificity of CNs.
 Specifically, the generated CN is designed to satisfy the following
requirements: (a) retain the retrieved counter-knowledge, (b) counter HS, and
(c) maintain fluency. Constraints (a) and (c) are implemented through a
differentiable n-gram matching function~\citep{liu2022don}, and the
cross-entropy between the token distribution of generated CNs and the neural
distribution of PLMs. For constraint (b), we
design a counter prompt to utilize parametric knowledge of PLM, ensuring that
the generated samples hold different opinions from HS. Then, we build a CN
classifier to predict the probability of the generated samples countering HS
and back-propagate with its cross-entropy loss.

In summary, the contributions of this study are as follows:
\begin{itemize}
\item This study designs a retrieval-augmented zero-shot CN generation method for HS, namely ReZG, to generate effective and specific CNs in the absence of in-target annotated corpora. ReZG integrates information retrieval and constrained decoding into CN generation, solving the difficulty of knowledge expansion and zero-shot generation.

\item To reduce \stmdocstextcolor{black}{the} non-countering
information in traditional semantic-related retrieval results, 
\stmdocstextcolor{black}{this study proposes a multi-dimension retrieval method based on the \textbf{S}tance consistency, \textbf{S}emantic overlap rate, and \textbf{F}itness for HS, referred to as SSF. The SSF method employs these diverse metrics to progressively extract finer-grained counter-knowledge layer by layer.}
Experimental results show that the SSF method outperforms both widely-used
  sparse retriever BM25~\citep{robertson2009probabilistic}, and
  dense retriever DPR~\citep{karpukhin2020dense} in the CN
  generation task.

\item This study proposes an energy-based generator for CNs, which utilizes
 knowledge preservation, countering, and fluency differentiable
  constraints instead of in-target CNs as control signals to fuse
  external knowledge and generate CNs in a zero-shot setting.  Experimental
  results show that on the CN generation task, the energy-based generator has a stronger ability to integrate external counter-knowledge than prompt-based PLM generators which have at least ten times larger parameters,
  such as Llama2-70B-chat~\citep{touvron2023Llama} and
  GPT-3.5-\stmdocstextcolor{black}{Turbo~\cite{gpt3}}.

\item We conducted a detailed evaluation of the generated CNs from multiple perspectives including relevance, success rate of countering (SROC), toxicity, effective persuasiveness, novelty, etc. 
Experimental results indicate ReZG has stronger generalization capabilities and outperforms state-of-the-art (SOTA) baselines in both automatic and human evaluations, achieving significant improvements of 2.0\%+ in relevance and 4.5\%+ in SROC metrics.
\end{itemize}

\section{Related works}

\subsection{\textcolor{black}{CN} collection}
In early research, CNs were mainly acquired by human annotation or
writing.~\citet{mathew2019thou} manually annotated and published the first CN
dataset and conducted a detailed psycholinguistic analysis based
\stmdocstextcolor{black}{thereon}. 
This dataset, which comprises 6,898 CNs and 7,026 non-CNs, was annotated from YouTube comments by two PhD\stmdocstextcolor{black}{s} and three undergraduates.
\stmdocstextcolor{black}{Subsequently},~\citet{qian2019benchmark} published two
large-scale HS intervention datasets, which provide conversation snippets
collected from Reddit\footnote{\url{https://www.reddit.com}.}
and Gab.\footnote{\url{https://gab.ai}.}
\stmdocstextcolor{black}{The datasets contain} approximately 40,000
intervention responses written by 926 mechanical Turk4 workers. Considering
that the same statement may have different meanings in different
contexts,~\citet{yu2022hate} explored the role of conversational context in
annotating and detecting HS\stmdocstextcolor{black}{s} and
CN\stmdocstextcolor{black}{s,} and released a context-aware dataset. They
invited hundreds of non-expert operators to annotate Reddit comments with and
without context, marking whether it was HS, CN, or neutral speech. Their
experiments also \stmdocstextcolor{black}{demonstrated} the effectiveness of
stance-based pre-training in detecting CNs.

\stmdocstextcolor{black}{T}he collection of CNs gradually
progress\stmdocstextcolor{black}{ed} from monolingual to multilingual and from
crowdsourcing to
\stmdocstextcolor{black}{n}ichesourcing.~\citet{chung2019conan} proposed the
first high-quality, multilingual, expert-based CN dataset, which contains
4,078 HS/CN pairs in English, French, or Italian.

To reduce the manual workload,~\citet{tekirouglu2020generating} implemented an
author-reviewer framework \stmdocstextcolor{black}{for collecting} CNs. In this
framework, GPT2~\citep{radford2019language} acts as the author to ``write'' CNs
and \stmdocstextcolor{black}{a human} serves as the reviewer to filter and
post-edit CNs.~\citet{fanton2021human} proposed a human-in-the-loop CN data
collection method, which also conducted expert-based filtering and post-editing
on the generated CNs. This method iteratively adds the post-edited samples to
the training data to fine-tune and optimize the generative language model. They
\stmdocstextcolor{black}{eventually} generated 5,000 HS/CN pairs, which took
\stmdocstextcolor{black}{three reviewers approximately 18 weeks}.
~\citet{bonaldi2022human} followed this author-reviewer
pipeline and specifically designed new strategies to collect CNs in multi-turn
dialogs.

\subsection{\textcolor{black}{CN} generation}
\stmdocstextcolor{black}{With the availability of these} expert-authored
datasets, many seminal studies \stmdocstextcolor{black}{have made} use of NLG
technology to automatically or semi-automatically generate CNs
\citep{tekirouglu2020generating,tekiroglu2022using}.
\citet{qian2019benchmark} evaluated the performance of several basic generative
models (e.g.~sequence-to-sequence model\stmdocstextcolor{black}{s}) in
generating CNs and found that much room still
remain\stmdocstextcolor{black}{ed} for improvement because generated CNs
\stmdocstextcolor{black}{were} sometimes irrelevant and mostly commonplace.
To address this lack of variety and relevance,~\citet{zhu2021generate} proposed
\stmdocstextcolor{black}{the} Generate, Prune, Select (GPS) three-module
pipeline method. \stmdocstextcolor{black}{This} method generates multiple
candidate samples to increase diversity and uses retrieval-based selection to
increase \stmdocstextcolor{black}{the} relevance.
Additionally,~\citet{chung2021towards} combined generative models with
information retrieval technology to incorporate external knowledge into the
target samples, aiming to promote the informativeness of CNs and avoid
hallucinatory phenomena.
These seminal studies \stmdocstextcolor{black}{included} training a supervised
model on expert-authored datasets with various strategies to enhance the
diversity, richness or reliability of CNs
~\citep{zhu2021generate,chung2021towards,tekiroglu2022using,gupta2023counterspeeches}.

\stmdocstextcolor{black}{Recent} studies have begun to explore generating CNs
with PLM\stmdocstextcolor{black}{s} in low-resource languages and
few-\stmdocstextcolor{black}{shot}
prompting~\citep{chung2020italian,ashida2022towards}, as PLMs have demonstrated
a strong few-shot ability in NLG tasks.
\citet{tekiroglu2022using} evaluated and compared the performance of several
advanced PLMs including BERT~\citep{kenton2019bert},
GPT-2~\citep{radford2019language}, and T5~\citep{raffel2020exploring} in
end-to-end supervised setting. They found that the autoregressive model
combined with stochastic decoding can achieve
\stmdocstextcolor{black}{superior} performance.
\citet{chung2020italian} and~\citet{ashida2022towards} explored generating CNs with PLMs in low-resource languages and few-shots
prompting\stmdocstextcolor{black}{, respectively}.

The above methods have achieved good performance in in-target CN generation. However, when dealing with HS of unseen targets, \stmdocstextcolor{black}{generating} specific and effective CNs \stmdocstextcolor{black}{is difficult owing} to the lack of new target-related information. Meanwhile, the \stmdocstextcolor{black}{challenge of} to directly retriev\stmdocstextcolor{black}{ing} and integrat\stmdocstextcolor{black}{ing} external counter-knowledge leaves the generated CNs with outdated evidence. As the annotation speed of CN data cannot keep up with the emergence of new HS targets, it is extremely important to study a zero-shot method \stmdocstextcolor{black}{for} generating CNs. Therefore, this \stmdocstextcolor{black}{study} combines information retrieval and constraint decoding technology to propose a retrieval-augmented zero-shot CN generation technology. The differences between our work and other works are summarized in Table \tabref{tab:realted works}.

\begin{table*}
\centering
\caption{The comparison of ReZG with other CN generation methods from the following three aspects: its ability to integrate external counter-knowledge, whether it is zero-shot, and its requirement for human participation.}
\begin{tabular}{c|c|c|c}
  \hline
  \textbf{Method} &\textbf{External counter-knowledge?}
  & \textbf{Zero-shot?}
  & \textbf{Human-free?}\\
  \hline
  Qian et al. \citep{qian2019benchmark}  & \ding{55}& \ding{55} & \ding{55} \\
  Chung et al. \citep{chung2020italian}    & \ding{55}& \ding{55} & \ding{51}\\
  Tekiroglu et al.  \citep{tekirouglu2020generating} & \ding{55}& \ding{55} & \ding{55}\\
  Zhu et al.\citep{zhu2021generate}   & \ding{55}& \ding{55} & \ding{51}\\
  Chung et al. \citep{chung2021towards} & \ding{51}& \ding{55} & \ding{51} \\
  Tekiroglu et al. \citep{tekiroglu2022using}    & \ding{55}& \ding{55} & \ding{51}\\
  Ashida et al. \citep{ashida2022towards} &  \ding{55}& \ding{55}& \ding{51}\\
  Gupta et al. \citep{gupta2023counterspeeches} &  \ding{55}& \ding{55}& \ding{51}\\
  \textbf{Ours }& \ding{51} & \ding{51}& \ding{51}\\
  \hline
\end{tabular}
\label{tab:realted works}
\end{table*}

\subsection{Retrieval-augmented generation}
Retrieval-Augmented Generation (RAG) is a technique that enhances the model's
ability to generate complex sequences by retrieving extra knowledge. Owing to
its interpretability, scalability, and adaptability, this paradigm has been
widely applied in knowledge-intensive generation tasks such as open-domain
question answering~\citep{siri2023improving}, code
generation~\citep{zhou23docprompting}, generative information
extraction~\citep{cong2023universal}, dialog generation~\citep{LIU2023126878}
and machine translation~\citep{ye2023tram}.

In the RAG pipeline, the retriever fetches supplementary knowledge from an
external knowledge base, and then the generator combines the retrieved
information with the parameterized knowledge learned during pre-training to
address the given task. Typically, retrievers employ sparse or dense similarity
measures to filter information. Sparse-vector retrievals such as TF--IDF and
BM25~\citep{robertson2009probabilistic} compute similarity by matching keywords
in the sparse bag-of-words representation space, while dense vector retrievals
determine similarity by calculating the inner product between low-dimensional
dense vectors obtained from neural
networks~\citep{karpukhin2020dense,lin2023aggretriever,wu2023contextual}.
Sparse-vector retrievals can check for precise term overlap, whereas dense
vector retrievals can retrieve information that is semantically related but may
differ in wording. Both are similarity-based retrievals.

However, the most similar generic information may not be the most suitable for
downstream models~\citep{li2022survey}, and the judgement of similarity also
depends on specific aspects~\citep{danqi2023csts}. Hence, some research has
focused on task-specific
retrieval~\citep{cai2021neural,ren2023retrieve,dai2023promptagator}, which
tailors and optimizes retrieval methods for specific task objectives. For
example,~\citet{ren2023retrieve} found that similarity-based retrieval
enhancement strategies were not applicable to document-level Event Argument
Extraction (EAE). Thus, they explored how to design the retrieval augmentation
strategy for document-level EAE considering input and label distribution and
proposed an adaptive hybrid retrieval augmentation strategy. Meanwhile,
PROMPTAGATOR~\citep{dai2023promptagator} utilized the large language model (LLM) as a
few-shot query generator to create task-specific retrievers. Like previous
work, we also customized a hierarchical retrieval method tailored for the task
of CN generation, to reduce non-countering
information in traditional semantic-related retrieval results.

 \begin{figure*}[htbp]
  \centering
  \includegraphics[width=1.0\textwidth]{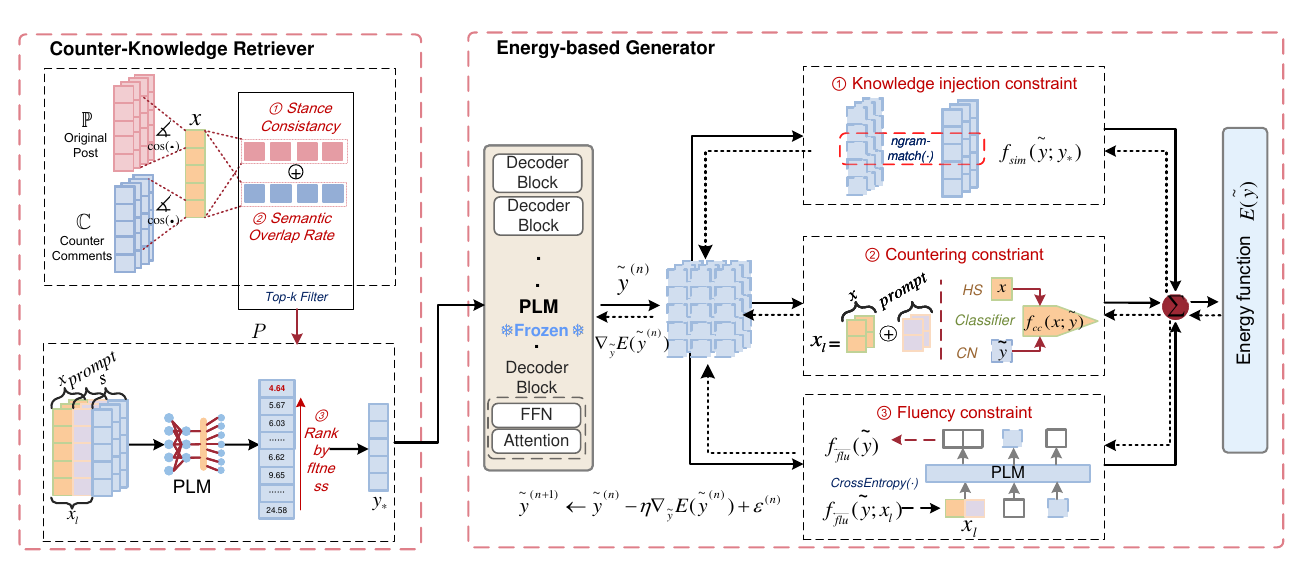}
 \caption{ReZG framework.
  $s$ denotes the sentence extracted from filtered counter\stmdocstextcolor{black}{-}comments. $x$, $y_{*}$, and $\widetilde{y}$ represent HS, retrieved counter-knowledge, and the continuous representation of generated CNs, respectively. The $f_{\operatorname{sim}}$($\cdot$), 
  $f_{\operatorname{cc}}$($\cdot$), and $f_{\overleftarrow{\overrightarrow{\mathrm{flu}}}}$($\cdot$) in the generator denote knowledge preservation, countering and fluency constraints, which are added to the energy function $E\left(\widetilde{y}\right)$.}
  \label{pic:framework}
\end{figure*}

\section{Methodology}
\begin{table}[ht]
\renewcommand{\arraystretch}{1.3}
\caption{Symbols and notations}
\label{tab:symbol}
\centering
\begin{tabular}{|p{1.1cm}||p{6.5cm}|}
\hline
\textbf{Symbol}& \textbf{Description}\\
\hline
\hline
$\mathbb{R}$ &  Set of real numbers \\

$\mathbb{P}$, $\mathbb{C}$, $\mathbb{S}$ &  Sets of posts, counter{-}comments and sentences, respectively \\

$p_{i}$, $c_{i}$, $s_{i}$ &  The $i$-{th} post, counter{-}comment and sentence\\

$P$ & Prompt used in the prompt learning \\

$x$  &  Hate speech \\

$y$  &  Target CN \\

$\widehat{c_{i}}$, $\widehat{x_{i}}$ & Semantic embeddings of $c_{i}$ and $x_{i}$ \\

 $\widetilde{p_{i}}$, $\widetilde{x_{i}}$ & Stance embeddings of $p_{i}$ and $x_{i}$\\
 
$\mathbf{x},\mathbf{y}$ &  Vector representations of $y$ and $x$ \\

$V$
& Set of the vocabulary\\

$\mid V\mid$ & The size of set $V$\\

$p$($\mathbf{y}_{\mathbf{i}}$) & Probability distribution of the $i$-{th} word $y_{i}$\\

$p\left(\cdot\mid\mathbf{y}_{<\mathbf{t}}\right)$ & Probability distribution of the next token with the input tokens $\mathbf{y}_{<t}$ \\

 $\tilde{\mathbf{y}}$ & Output logits of $y$ from PLM \\

$\tau$ & Temperature parameter \\

$W$, $b$ & Matrices and bias vectors \\
\hline
\end{tabular}
\end{table}

ReZG consists of two modules, a \textbf{C}ounter-\textbf{K}nowledge \textbf{R}etriever (CKR) and \stmdocstextcolor{black}{an} \textbf{En}ergy-based \textbf{G}enerator for CNs (EnG), as depicted in Fig.\ref{pic:framework}. The CKR is \stmdocstextcolor{black}{used} for acquiring information containing counter-knowledge from \stmdocstextcolor{black}{an} external knowledge repository, and EnG is responsible for generating CNs based on the context and retrieved information. These two modules are described in detail in the following Sections \ref{sec:retrieve} and   \ref{sec:eng}, \stmdocstextcolor{black}{respectively}, and the main symbols and notation\stmdocstextcolor{black}{s} used in this paper are displayed in Table \tabref{tab:symbol}.

\subsection{Counter-knowledge retriever}
\label{sec:retrieve}
In the field of dialog systems, post-dialog can be obtained by compiling the historical responses of semantically similar dialogs. By analogy, the counter-knowledge in historical responses to the posts that have stances similar to HS can be leveraged to support the generation of corresponding CNs. Following this paradigm, we first construct
 a knowledge repository with counter-structures by gathering a substantial corpus of HS and its historical counter
 comments from the ChangeMyView(CMV)\footnote{\url{https://www.reddit.com/r/changemyview/wiki/index/}.} forum.
\stmdocstextcolor{black}{Then, a hierarchical retrieval method based on the \textbf{S}tance consistency, \textbf{S}emantic overlap rate, and \textbf{F}itness for HS, referred to as SSF,} is proposed to retrieve fine-grained counter-knowledge layer by layer from posts to comments and further to more specific sentences. 
Details of the constructed knowledge repository and SSF retrieval method are elaborated below.

\paragraph*{\textbf{Knowledge repository {c}onstruction}}
\label{sec:ck}

We use \stmdocstextcolor{black}{the} CMV subreddit as the source of external knowledge, as it provides a wealth of positive counter-knowledge. \stmdocstextcolor{black}{The} CMV subreddit is a forum where people can publish original posts to express their views and \stmdocstextcolor{black}{challenge other views} through commenting in the discussion \stmdocstextcolor{black}{in an effort} to change \stmdocstextcolor{black}{the} opinion holder's mind. In CMV, the poster can issue a delta/$\triangle$ to the comments that change its mind, and bystanders can express support and opposition to the counter\stmdocstextcolor{black}{-}comments by giving up\stmdocstextcolor{black}{-vote}s and down-\stmdocstextcolor{black}{vote}s. This provides a quantifiable metric for evaluating whether a review is persuasive. The targets of HS are used as queries to retrieve related posts in CMV. We reserve \stmdocstextcolor{black}{the} top 100 posts for each query. \stmdocstextcolor{black}{Subsequently}, to ensure the quality and persuasiveness of the comments, \stmdocstextcolor{black}{those with} fewer up\stmdocstextcolor{black}{-votes} than down\stmdocstextcolor{black}{-vote}s are deleted. We denote this knowledge repository with counter structures as <$ \mathbb{P},\mathbb{C}$>, where $\mathbb{P}$ \stmdocstextcolor{black}{represents} a series of posts that articulate viewpoints and $\mathbb{C}$ are \stmdocstextcolor{black}{the} corresponding counter\stmdocstextcolor{black}{-}comments. \stmdocstextcolor{black}{The knowledge repository} eventually comprises approximate\stmdocstextcolor{black}{ly} 2,800 posts and 63,000 counter\stmdocstextcolor{black}{-}comments.

 \begin{figure*}[htbp]
  \centering
  \includegraphics[width=.88\textwidth]{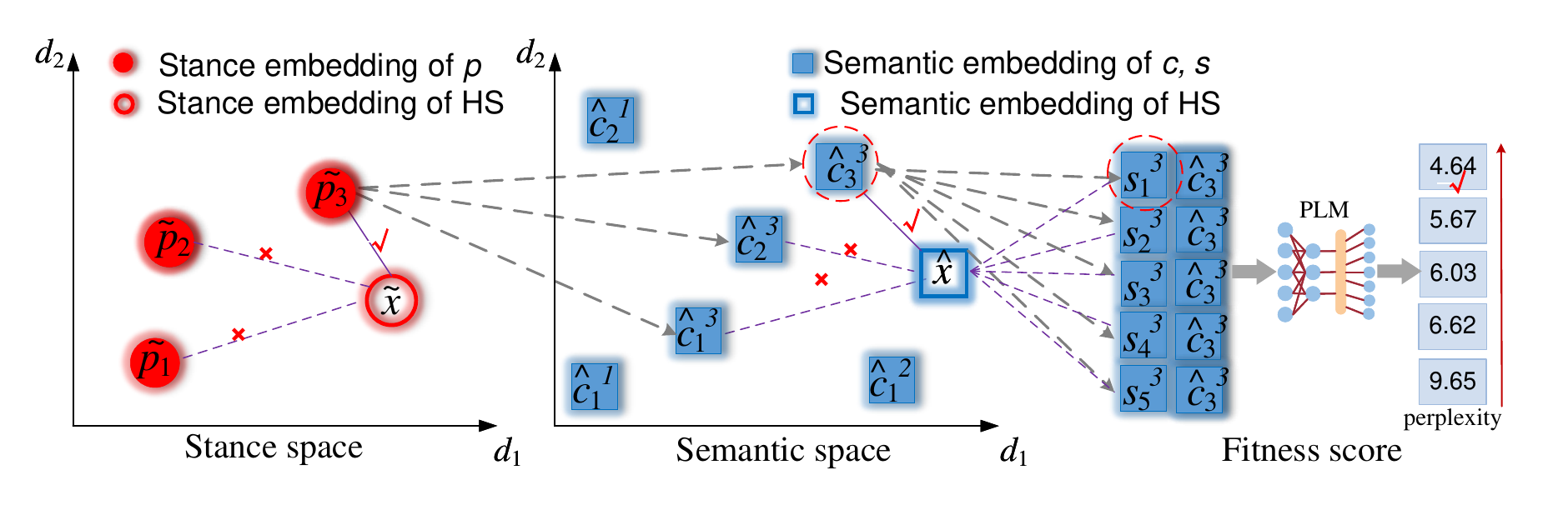}
  \caption{\textcolor{black}{An Illustration of information selection process in SSF Algorithm.}}
  \label{pic:ssf-process}
\end{figure*}
\paragraph{\textbf{Counter-knowledge retrieval module.}}

As shown in Algorithm \ref{alg:3step}, the SSF retrieval method includes three steps.
First, we filter the posts that exhibit similar stances with the given HS based on the stance consistency metric \textsl{STA}. 
\stmdocstextcolor{black}{Subsequently,} combined with the semantic similarity metric \textsl{SEM}, the counter\stmdocstextcolor{black}{-}comments of selected posts, that involve more \stmdocstextcolor{black}{of} the content discussed in \stmdocstextcolor{black}{the} HS and narrative contrarily around it, are retained.
Finally, through a constructed counter prompt, more fine-grained counter information is further extracted from the filtered comments from the perspective of language modeling based on perplexity.
\textsl{STA}($\cdot$), \textsl{SEM}($\cdot$), and \textsl{FIT}($\cdot$) in Algorithm \ref{alg:3step}{1} denote \stmdocstextcolor{black}{the} calculation formulas \stmdocstextcolor{black}{for} the stance consistency, semantic overlap rate, and fitness metrics used in the retrieval.

\textbf{Stance consistency.} As shown in the information selection process of the SSF method in  \ref{pic:ssf-process}, the cosine distance between \stmdocstextcolor{black}{the} posts and HS in the high-dimensional space representing the stance is used to measure \textsl{STA} between \stmdocstextcolor{black}{the} posts and HS. 
This criterion serves as a filter to identify posts that are most congruent with the stance of \stmdocstextcolor{black}{the} given HS, calculated as follows:
\begin{equation}
\begin{aligned}
\textsl{STA}\left(p_{i},x_{i}\right)=\frac{\widetilde{p_{i}}^{\top} \widetilde{x_{i}}}{\left\|\widetilde{p_{i}}\right\| \cdot\left\| \widetilde{x_{i}}\right\|}.
\label{eq:sta}
\end{aligned}
\end{equation}

$\widetilde{p_{i}}$ and $\widetilde{x_{i}}$ are the stance embeddings of the $i$-th post $p_{i}$ and HS $x_{i}$, $\top$ represents the transpose operation, and $\| \| $ represents the Euclidean norm.
The representation of stance in vector space is obtained by fine-tuning the pre-trained sentence similarity model using contrastive loss and stance detection datasets. The detailed acquisition method is \stmdocstextcolor{black}{described} in Appendix. \ref{sec:stance}.
The top-$k_{1}$ posts with \stmdocstextcolor{black}{the} highest STA are saved.

\begin{algorithm}[htb]
\renewcommand{\algorithmicrequire}{\textbf{Input:} }
\renewcommand{\algorithmicensure}{\textbf{Output:} }
\caption{SSF Retrieval Method for Counter-Knowledge}
\label{alg:3step}
\begin{algorithmic}[1]
    \REQUIRE Hate speech $x$ and a knowledge repository with counter-structure <$\mathbb{P},\mathbb{C}$>
    \ENSURE Counter-knowledge $y_{*}$ for HS
    \renewcommand{\algorithmicrequire}{\textbf{Return }}
        \STATE $L$=[] 
    \FOR{$p_i$ in $\mathbb{P}$} 
    \STATE $l_i$=$\text{\textsl{STA}}\left(p_i,x\right)$ \texttt{// Calculate the stance consistency with Eq.\ref{eq:sta}} \\
        $L$.append( \{$p_i$: $l_i$\} )
    \ENDFOR 
    \STATE $L_{sorted}$=sort$\left(L,\geq_{l_{i}}\right)$ 
    \STATE $\mathbb{P}_{k_{1}}$ = top$_{p} \left(L_{sorted},k_1\right)$
    \STATE $L=[]$
    \FOR{$p_{i}, c_{i},$ in $\mathbb{P}_{k_{1}},\mathbb{C}_{k_{1}}$ }
    \STATE $l_{i}$=$\alpha*\text{\textsl{STA}}\left(p_i,x\right)$+$\beta*\text{\textsl{SEM}}\left(c_i,x\right)$ \texttt{// Calculate the semantic overlap rate with Eq.\ref{eq4}} \\
     $L$.append( \{$c_i$: $l_i$\} )
    \ENDFOR
    \STATE $\mathbb{C}_{k_{2}}$ = top$_{c} \left(\text{sort}\left(L,\geq_{l_{i}}\right) ,k_2\right)$ 
    
    \STATE \textmd{Split $\mathbb{C}_{k_{2}}$ into a set of sentences $\mathbb{S}$} and  $L$=[] 
    \FOR{$s_i$ in $\mathbb{S}$} 
    \STATE $l_i$=$\text{\textsl{FIT}}\left(s_i,x\right)$ \texttt{// Calculate the fitness with Eq.\ref{eq5}} \\
        $L$.append( \{$s_i$: $l_i$\} )
    \ENDFOR 
    \STATE $y_{*}$ = top$_{s} \left(\text{sort}\left(L,\leq_{l_{i}}\right),k_3\right)$

    \REQUIRE  $y_{*}$
\end{algorithmic}
\end{algorithm}

\textbf{Semantic overlap rate.} We measure
\textsl{SEM} of \stmdocstextcolor{black}{the}
counter\stmdocstextcolor{black}{-}   comments and HS by calculating the cosine
distance between them in the semantic space, as shown in \eqref{eq4}.
Considering \stmdocstextcolor{black}{that} the contents used to counter HS
should not only counter \stmdocstextcolor{black}{the HS} but also have strong
pertinence to avoid general replies~\citep{zhu2021generate,zheng2023makes}, we
combine \textsl{SEM} and \textsl{STA} to filter out
counter\stmdocstextcolor{black}{-}comments that involve more about the
content\stmdocstextcolor{black}{s} discussed in HS and narrative contrarily
around it, as follows:
\begin{equation}
\begin{aligned}
\chi\left(c_{i},x_{i}\right)=\alpha*\textsl{SEM}\left(c_{i},x_{i}\right) +\beta*\textsl{STA}\left(p_{i},x_{i}\right),\\
\textsl{SEM}\left(c_{i},x_{i}\right)=\frac{\widehat{c_{i}}^{\top} \widehat{x_{i}}}{\left\| \widehat{c_{i}}\right\|  \cdot\left\|  \widehat{x_{i}}\right\| }\quad \quad \quad,
\end{aligned}\label{eq4}
\end{equation}
where $\widehat{c_{i}}$ and $\widehat{x_{i}}$ 
 are semantic embeddings of \stmdocstextcolor{black}{the} $i$-th comments $c_{i}$ and HS $x_{i} $ from the pre-trained sentence similarity model \textsl{sup-simcse-bert-base-uncased}.\footnote{\url{https://huggingface.co/princeton-nlp/sup-simcse-bert-base-uncased/}.} 
$\alpha$ and $\beta$ are weights for each metric.
The counter\stmdocstextcolor{black}{-}comments of saved posts are scored by $\chi\left(c_{i},x_{i}\right)$ and the top-$k_{2}$ ones are saved.

\textbf{Fitness.} Finally, to acquire finer-grained and more accurate    counter-knowledge, we \stmdocstextcolor{black}{design} a fitness function to determine which sentence $s$ in the filtered comments contains the counter-knowledge that best matches a specific HS, from the perspective of language modeling. $\textsl{FIT}$($\cdot$) is calculated using perplexity as follows:
\begin{equation}
\begin{aligned}
\textsl{FIT}\left(s_{i},x_{i}\right)=
e^{\mathcal{L}_{CE} \left( PLM\left(x_{i}\oplus P \oplus s_{i}\right) \right)},
\end{aligned}
\label{eq5}
\end{equation}
where $P$  $=$ [However, I disagree.], representing the counter prompt. $\oplus$ denotes the concatenation operation and $\mathcal{L}_{CE}$($\cdot$) denotes the cross-entropy loss of PLM with $x_{i}\oplus P \oplus s_{i}$ as input. The final results are sorted in ascending order and the top-$k_{3}$ results are selected as the counter-knowledge $y_{*}$ to be used in the EnG.

\subsection{Energy-based \textcolor{black}{g}enerator for CNs}
\label{sec:eng}

Although \stmdocstextcolor{black}{the} CKR \stmdocstextcolor{black}{acquires} the information that contains counter-knowledge, it still cannot be directly applied to counter the given HS. Meanwhile, the absence of training data makes it extremely difficult for the model build mappings from counter-knowledge to CNs.
Inspired by COLD decoding~\citep{qin2022cold}, decoding can be formulated as
sampling from an energy-based model~\citep{lecun2006tutorial} using Langevin
\stmdocstextcolor{black}{d}ynamics. In this \stmdocstextcolor{black}{manner},
control signals can be passed to the model via the energy function. Therefore,
we design different differentiable constraint functions and add them to the
energy function instead of CN references to address this problem.

The sampling process is formulated using the gradient
$\nabla_{\tilde{\mathbf{y}}} E\left(\tilde{\mathbf{y}}^{\text{(}n\text{)}}\right)$ as follows:
\begin{equation}
\tilde{\mathbf{y}}^{\text{(}n+1\text{)}} \leftarrow \tilde{\mathbf{y}}^{\text{(}n\text{)}}
-\eta \nabla_{\tilde{\mathbf{y}}} E\left(\tilde{\mathbf{y}}^{\text{(}n\text{)}}\right)
+\epsilon^{\text{(}n\text{)}},
\label{eq:y}
\end{equation}
where $E$($\cdot$) denotes the energy function, \stmdocstextcolor{black}{and} $\eta>$~0 and $\epsilon^{\text{(}n\text{)}} \in \mathcal{N}\left(0, \sigma\right)$ denote the step size and noise at the $n$-${th}$ iteration, respectively. $\tilde{\mathbf{y}}$ denotes the continuous representation of the generated CN.

\stmdocstextcolor{black}{First}, $\tilde{\mathbf{y}}$ is initialized using the retrieved counter-knowledge  $y_{*}$. \stmdocstextcolor{black}{As the} tokens of $y_{*}$ are discrete, discontinuous, and undifferentiated, we utilize the output logits $\tilde{\mathbf{y}}=[\tilde{\mathbf{y}}_{1}; \ldots; \tilde{\mathbf{y}}_{T}]$ of $y_{*}$ from the PLM as its representation, where $\tilde{\mathbf{y}}_{t} \in \mathbb{R}^{\mid V\mid}$, and $\mid V\mid$ and $T$ denote the vocabulary size and CN length, respectively.
\stmdocstextcolor{black}{Following} multiple iterations, every output token $\mathbf{y}_{i}$ is realized by sampling from $p\left(\mathbf{y}_{i}\right) \sim \sigma\left(\tilde{\mathbf{y}}_{i} \diagup
 \tau\right)$, where $\sigma$ represents the $\operatorname{softmax}$ operation and $\tau$ is the temperature parameter. When $\tau \rightarrow 0$, $p\left(\mathbf{y}_{i}\right)$ \stmdocstextcolor{black}{returns} to a one-hot vector, representing a discrete token.

 To meet the constraints of \stmdocstextcolor{black}{the} generated samples, \stmdocstextcolor{black}{namely} (a) retaining counter-knowledge\stmdocstextcolor{black}{,} (b) countering HS\stmdocstextcolor{black}{, and} (c) maintaining fluency, we transform these three constraints into the differentiable functions $f\left(\tilde{\mathbf{y}}\right)$ and incorporate them into the energy function $E\left(\tilde{\mathbf{y}}\right)$, as described in Section \ref{sec:constraint}.

\subsubsection{Constraint formulation}
\label{sec:constraint}
This subsection introduces the formulation of the aforementioned three constraints and the energy function.

\textbf{For constraint (a)}, we utilize the differentiable n-gram matching
function $\text{ngram-match}\left(\cdot,\cdot\right)$~\citep{liu2022don} to encourage $\tilde{\mathbf{y}}$ to have more
\stmdocstextcolor{black}{overlapping} tokens with counter-knowledge
$y_{*}$.  The \textit{knowledge preservation constraint} $f_{\operatorname{sim}}\left(\tilde{\mathbf{y}}\right)$ is
formulated as follows:
\begin{equation}
f_{\operatorname{sim}}\left(\tilde{\mathbf{y}};y_{\ast}\right)=\text{ngram-match}\left(\tilde{\mathbf{y}}, y_{\ast}\right);
\end{equation}

\textbf{For constraint (b)}, we first utilize a counter prompt that applies the implicit knowledge of \stmdocstextcolor{black}{the} PLM to make the generated \stmdocstextcolor{black}{CN} hold opinions \stmdocstextcolor{black}{that differ} from \stmdocstextcolor{black}{the} HS, as shown in Eq.~\ref{eq:pfluent}. Next, we use a CN classifier $p_{\phi}$ to evaluate whether the generated CN can counter the corresponding HS, and develop \textit{the countering constraint} $f_{\mathrm{cc}}$($\cdot$) with the cross-entropy of the logits of $p_{\phi}$\stmdocstextcolor{black}{,} as follows:
\begin{equation}
f_{\mathrm{cc}}\left(\mathbf{x};\tilde{\mathbf{y}}\right)=
\sum_{i=1}^{n}q_{\phi}\left(\mathbf{x},\tilde{\mathbf{y}}\right)_{i}\log \sigma\left(p_{\phi}\left(\mathbf{x},\tilde{\mathbf{y}}\right)_{i}\right) \Big/
\gamma,
\end{equation}
where $\mathbf{x}\in \mathbb{R}^{L*V} $ represents the one-hot vector of \stmdocstextcolor{black}{the} HS and $L$ denotes the length of \stmdocstextcolor{black}{the} HS.  $p_{\phi}\left(\mathbf{x},\tilde{\mathbf{y}}\right)$ represents the logits of the classifier $p_{\phi}$ with $\mathbf{x}$ and $\tilde{\mathbf{y}}$ as inputs, $q_{\phi}\left(\mathbf{x},\tilde{\mathbf{y}}\right)$ is the expected label (0, 1) for $\left(\mathbf{x},\tilde{\mathbf{y}}\right)$ pairs, and $n$ is the number of classes.
$\gamma$ is a scale factor for $f_{\mathrm{cc}}$($\cdot$). It is used to make the value of $f_{\mathrm{cc}}$($\cdot$) have the same order of magnitude \stmdocstextcolor{black}{as the} other two constraints.
Specifically, $p_{\phi}$ is acquired by fine-tuning
\stmdocstextcolor{black}{the} PLM with an additional linear layer on CN
datasets~\citep{chung2019conan,qian2019benchmark,chung2021towards}. $p_{\phi}$ reaches an
accuracy of 90.60\%. It is calculated as follows:
\begin{equation}
\begin{aligned}
p_{\phi}\left(\mathbf{x},\tilde{\mathbf{y}}\right)= PLM& \left( \operatorname{join}\left(\mathbf{x},\tilde{\mathbf{y}}\right)\right)*W^{T}+b,\\
\operatorname{join} \left(\mathbf{x}, \tilde{\mathbf{y}}\right) =
\text{[BOS] } & \mathbf{x} \text{ [EOS] }\sigma\left(\tilde{\mathbf{y}}\Big/ \tau\right) \text{ [EOS]},
\end{aligned}
\label{eq:join}
\end{equation}
where $W \in \mathbb{R}^{2 \times d}$  and $b \in \mathbb{R}^{2}$ denote learnable matrices and bias vectors respectively; [EOS] is the end token of \stmdocstextcolor{black}{the} PLM, \stmdocstextcolor{black}{which is} used to separate $\mathbf{x}$ and $\tilde{\mathbf{y}}$; \stmdocstextcolor{black}{and} [BOS] is the beginning token of PLM, \stmdocstextcolor{black}{which is } used as a prefix token to obtain contextual embedding for the entire sequence.
The pair of $\mathbf{x}$ and $\tilde{\mathbf{y}}$ \stmdocstextcolor{black}{is} concatenated into one sequence as input of the classier as shown in Eq.~\ref{eq:join}. The $\operatorname{softmax}$ operation is performed on $\tilde{\mathbf{y}}$ to restore it to the one-hot vector before the concatenation because the input of the classier should be a set of discrete tokens.

\textbf{For constraint (c)}, it consists of a 
\textit{left-to-right fluency constraint} $f_{lr}$($\cdot$)
that encourages $\tilde{\mathbf{y}}$ to be fluent and coherent with the left-side context, and a \textit{right-to-left fluency constraint }
$f_{rl}$ ($\cdot$) 
that encourages  $\tilde{\mathbf{y}}$ to be coherent with the right-side context. \stmdocstextcolor{black}{The} \textit{fluency constraint} $f_{\operatorname{flu}}\left(\tilde{\mathbf{y}}\right)$ is formulated as follows:
\begin{equation}
\begin{aligned}
&\mathbf{x}_{l}=\mathbf{x}\oplus P,\\
&f_{rl}\left(\tilde{\mathbf{y}}\right)=\sum_{t=1}^{T}\sum_{v\in V}p_{\overleftarrow{\mathrm{PLM}}}\left(v\mid\tilde{\mathbf{y}}_{>t}\right)* \operatorname{log} \sigma\left(\tilde{\mathbf{y}}_{t}\text{(}v\text{)}\right),\\
&f_{lr}  \left(\tilde{\mathbf{y}}; \mathbf{x}_{l}\right)= 
 \sum_{t=1}^{T}\sum_{v\in V}  p_{\overrightarrow{\mathrm{PLM}}}\left(v\mid\tilde{\mathbf{y}}_{<t},\mathbf{x}_{l}\right)*
\operatorname{log}\sigma\left(\tilde{\mathbf{y}}_{t}\text{(}v\text{)}\right),\\
&f_{\operatorname{flu}}\left(\tilde{\mathbf{y}}\right) = \lambda_{c}^{l r}f_{lr} \left(\tilde{\mathbf{y}}; \mathbf{x}_{l}\right) + \lambda_{c}^{r l} f_{rl}\left(\tilde{\mathbf{y}}\right).
\end{aligned}
\label{eq:pfluent}
\end{equation}

$p_{\overrightarrow{\mathrm{PLM}}}\left(\cdot\mid\tilde{\mathbf{y}}_{<{\mathbf{t}}}\right)$ denotes the probability distribution of the next token when feeding \stmdocstextcolor{black}{the} PLM with the tokens $\tilde{\mathbf{y}}_{<t}$.
$\mathbf{x}_{l}$ is the left-side context for conditions \stmdocstextcolor{black}{of} the generation. It is a concatenation of $\mathbf{x}$ and the counter prompt $P$, encouraging $\tilde{\mathbf{y}}$ to be fluent and coherent with the counter of \stmdocstextcolor{black}{the} HS. The hyper-parameters $\lambda_{c}^{l r}$ and $\lambda_{c}^{r l}$ denote the weights of the corresponding constraint functions.

\textbf{Energy \stmdocstextcolor{black}{f}unction: }Finally, the weighted sum of the constraint functions $f_{\operatorname{sim}}\left(\tilde{\mathbf{y}}; \mathbf{y}_{\ast}\right)$, $f_{\mathrm{cc}}\left(\mathbf{x};\tilde{\mathbf{y}}\right)$ and
$f_{\operatorname{flu}}\left(\tilde{\mathbf{y}}\right)$ constitute the energy function $E\left(\tilde{\mathbf{y}}\right)$ as follows:
\begin{equation}
\begin{split}
E\left(\tilde{\mathbf{y}}\right)= \lambda_{a} f_{\operatorname{sim}}\left(\tilde{\mathbf{y}}; \mathbf{y}_{\ast}\right) +\lambda_{b} f_{\mathrm{cc}}\left(\mathbf{x};\tilde{\mathbf{y}}\right)+ \lambda_{c}f_{\operatorname{flu}}\left(\tilde{\mathbf{y}}\right),
\end{split}\label{eq:energy}
\end{equation}
where $\lambda_{a}$, $\lambda_{b}$ and $\lambda_{c}$ are hyperparameters that represent the weight\stmdocstextcolor{black}{s} of the corresponding constraint function\stmdocstextcolor{black}{s}.

\section{Experimental setup}
\subsection{Datasets}
Following~\citet{tekiroglu2022using}, we conduct experiments on the
Multitarget-CONAN dataset~\citep{fanton2021human}, as it is the only available
multi-target expert-authored dataset.
 It includes 5,003 <$HS, CN$> pairs and 3,709 HS samples, divided into an expert-collected seed subset $V_{1}$ and five generate-then-edit subsets $V_{2-6}$. The subset $V_{1}$, \stmdocstextcolor{black}{encompassing 880 <$HS, CN$>  pairs and 352 HS samples, is used as the test set}.
 The HS samples \stmdocstextcolor{black}{include} eight targets: DISABLED, JEWS, LGBT+, MIGRANTS, MUSLIMS, POC, WOMEN, and OTHER.

\subsection{Hyperparameters}

Following the basic configurations of COLD decoding~\citep{qin2022cold}, we
utilize GPT2-XL~\citep{radford2019language} as our base model, and the number
of sampling iterations $N$ and the step size $\eta$ are  set to
2,000 and 0.1 (Eq.~\ref{eq:y})\stmdocstextcolor{black}{, respectively}.
The ratio of left-to-right fluency 
to right-to-left fluency in the energy function is 9:1.
The knowledge preservation, countering and fluency constraints, are balanced with a 1:2:1 ratio, leading to $\lambda_{a}=0.25$, $\lambda_{b}=0.5$,
$\lambda_{c}^{l r}=0.225$ and $\lambda_{c}^{r l}=0.025$ \eqref{eq:pfluent,eq:energy}.
We set $k_{1},k_{2}$ and $k_{3}$ in the retriever model to 30, 10, and 10 respectively, and the \stmdocstextcolor{black}{maximum} length of generated samples is set to 30. We use \stmdocstextcolor{black}{the} PyTorch framework\footnote{\url{http://pytorch.org/docs/master/index.html}.} to implement our model and \stmdocstextcolor{black}{conduct} experiments on two NVIDIA A40 GPUs.

\subsection{Baselines}
We compare our method with three strong supervised methods and eight \stmdocstextcolor{black}{zero-shot} methods.

\textbf{Supervised baselines.} We \stmdocstextcolor{black}{select the} following three supervised methods, which have \stmdocstextcolor{black}{exhibited} outstanding performance \stmdocstextcolor{black}{in} \stmdocstextcolor{black}{CN} generation, as our baselines.
\begin{itemize}
   \item \textbf{GPS}~\citep{zhu2021generate} (Generate, Prune, Select) is a
    novel three-module    pipeline approach for generating diverse and
  relevant CNs. It utilizes \stmdocstextcolor{black}{a}
  variational autoencoder to generate a large pool of diverse
  response candidates, prunes ungrammatical candidates, and
  selects the most relevant one as the final CN.
   \item \textbf{GPT2-ft}~\citep{tekiroglu2022using} and
    \textbf{DialoGPT-ft}~\citep{tekiroglu2022using} are obtained by
  fine-tuning \stmdocstextcolor{black}{the}
  GPT2~\citep{radford2019language} and
  DialoGPT~\citep{zhang2020dialogpt}
  models with
  expert-authored CN datasets. ~\citet{tekiroglu2022using}
  conduct\stmdocstextcolor{black}{ed} an extensive study on
  fine-tuning various powerful PLMs to complete the CN generation
  task, and found that the fine-tuned GPT2 and DialogGPT can
  achieve the best performance with stochastic decoding. Thus, we
  \stmdocstextcolor{black}{select these models} as the supervised
  baselines.
\end{itemize}

\textbf{Zero-shot baselines.}
Since we \stmdocstextcolor{black}{could} not find \stmdocstextcolor{black}{a zero-shot} method for generating CNs for \stmdocstextcolor{black}{HSs}, we employ the prompt learning \stmdocstextcolor{black}{in} the generative PLM\stmdocstextcolor{black}{s} with \stmdocstextcolor{black}{a} zero-shot ability as our baselines. Two advanced generative PLM\stmdocstextcolor{black}{s} with different decoding mechanisms are \stmdocstextcolor{black}{selected} as our baselines.
One is the Transformer-based decoder language model,
\emph{GPT2-XL}~\citep{radford2019language}, and the other is the
Transformer-based encoder-decoder language model,
\emph{T0}~\citep{sanh2022multitask}, which is
\stmdocstextcolor{black}{approximately} 8 times larger than GPT2-XL and
demonstrates stronger zero-shot ability than the LLM GPT-3 on some natural language inference and sentence completion tasks.

To \stmdocstextcolor{black}{ensure} fairness \stmdocstextcolor{black}{in} the experiments, we use the same counter prompt as that in the ReZG method to guide GPT2 and T0 to generate CNs. Meanwhile, two stochastic decoding mechanisms (\stmdocstextcolor{black}{t}op-$k$ and \stmdocstextcolor{black}{t}op-$P$) and two deterministic decoding mechanisms (\stmdocstextcolor{black}{b}eam \stmdocstextcolor{black}{s}earch and \stmdocstextcolor{black}{g}reedy \stmdocstextcolor{black}{s}earch) are applied to the sampling stage of both GPT2-XL and T0. Accordingly, GPT2-XL and T0 with \stmdocstextcolor{black}{the} above four decoding mechanisms are respectively denoted as:
\textbf{GPT2-P}$_{topk}$, \textbf{GPT2-P}$_{topp}$, \textbf{GPT2-P}$_{bs}$, \textbf{GPT2-P}$_{gs}$,\textbf{T0-P}$_{topk}$,\textbf{T0-P}$_{topp}$,\textbf{T0-P}$_{bs}$, \stmdocstextcolor{black}{and}
\textbf{T0-P}$_{gs}$. $topk,topp,bs$ and $gs$ represent Top-$k$, Top-$\mathbb{P}$, \stmdocstextcolor{black}{b}eam \stmdocstextcolor{black}{s}earch, \stmdocstextcolor{black}{g}reedy \stmdocstextcolor{black}{s}earch decoding mechanisms respectively.

\subsection{Evaluation metrics}
\label{sec:evalu}
\subsubsection{Automatic evaluation}
\stmdocstextcolor{black}{As a} valid CN with high specificity should be highly
relevant to HS and counter it, we adopt the \textbf{relevance (Rel.)} and
\textbf{ success rate of countering (SROC)} metrics to evaluate the
effectiveness of CNs against HS. In addition,
following~\citet{chung2021towards} and \citet{tekiroglu2022using}, we assess the text
quality of generated CNs from \stmdocstextcolor{black}{the} following
\stmdocstextcolor{black}{five} perspectives: \textbf{toxicity (Tox.),
persuasiveness (Pers.), informativeness (Inf.), novelty (Nov.)} and
\textbf{linguistic quality (LQ)}. 
These metrics are automatically evaluated by the following methods:
\begin{itemize}
  \item \textbf{Relevance} between CN and HS is evaluated by
   BM25~\citep{robertson2009probabilistic}, which is an extensively used
  relevance metric in information retrieval.
  \item For \textbf{SROC} metric, we implement a CN \stmdocstextcolor{black}{classifier}\footnote{\stmdocstextcolor{black}{The resource of the CN classifier is available at: \url{https://github.com/MollyShuu/ReZG}}.}
  with an accuracy of 92.8\% to evaluate whether a generated CN is a specified
  and successful one that counters HS. The proportion of
  generated CNs with successful countering over all generated CNs
  is used to represent the SROC. This classifier is obtained by
  fine-tuning the pre-trained Roberta~\citep{liu2019roberta}
  model on the  CONAN~\citep{chung2019conan},
  Multitarget-CONAN~\citep{fanton2021human} and
  multitarget\_KN\_grounded\_CN~\citep{chung2021towards}
  datasets. The CNs in these datasets are expert-based and own a
  good specificity.
  
  \item \textbf{Toxicity} indicates the degree of disrespect, rudeness and
   unreasonableness~\citep{tekiroglu2022using}. Since toxicity content
  will pollute the online environment and intensify hatred, it
  needs to be evaluated and avoided. The toxicity is evaluated via
  the Perspective
  API\footnote{\url{https://www.perspectiveapi.com/}.} created
  by Google's Counter Abuse Technology team and Jigsaw, which has
  been used widely and commercially deployed in detecting
  toxicity
  \citep{si2022so,gehman2020realtoxicityprompts,yang2023harnessing}.

  \item \textbf{Persuasiveness} and
   \textbf{\stmdocstextcolor{black}{i}nformativeness}  of a sample are
  evaluated by means of advanced
  GPTScore~\citep{fu2024gptscore} with GPT-3.5-turbo 
  as the evaluator. GPTScore is a novel evaluation framework \stmdocstextcolor{black}{that} can effectively allow us to achieve \stmdocstextcolor{black}{the desired evaluation of} texts simply \stmdocstextcolor{black}{using} natural language instructions. \stmdocstextcolor{black}{Furthermore}, GPT-3.5-turbo as one of the most powerful LLMs, 
  \stmdocstextcolor{black}{has been} proven to rate text like human experts and
  can explain its own decision well~\citep{chiang2023can}.
  \stmdocstextcolor{black}{Among all the GPT-3.5 Turbo models, we
  select the latest version as of October 2024,
  gpt-3.5-turbo-0125,\footnote{\stmdocstextcolor{black}{\url{https://platform.openai.com/docs/models/gpt-3-5-turbo}}. (In the previous preprint version, it was gpt-3.5-turbo-0613 that was used to perform a batch evaluation on multiple samples.)}
  as the evaluator due to its superior accuracy at responding in
  requested formats. Subsequently, we design an evaluation prompt
  as shown in the box below 
  to make \stmdocstextcolor{black}{GPT-3.5-Turbo}
 score a sample from 0 to 1 in the aspects of persuasiveness and informativeness. Every sample is scored independently to avoid the inter-sample influence.}

\begin{center}
\fcolorbox{black}{gray!10}{\parbox{.95\linewidth}{
\hspace{\fill} \textbf{Evaluation Prompt}\hspace{\fill} 

You are now a language evaluation model.

Your task is to assess the following sample on a scale from 0 to 1, using the following criteria:

Persuasiveness: The degree to which the text effectively convinces or influences the reader's beliefs, attitudes, or opinions. A score of 0 indicates no persuasive elements, while a score of 1 implies highly compelling and convincing language.

Informativeness: The richness and depth of information conveyed in the text. A score of 0 suggests a lack of valuable information, while a score of 1 signifies a highly informative and comprehensive piece of writing.

Please provide the Persuasiveness and Informativeness scores of the following sample in JSON format without explanation. 

Here is the sample:
}}
\end{center}
  
Given that only the persuasiveness/informativeness of samples that refute rather than support HS are effective, we use the average score of the product of the CN classifier's results and the GPTScore to represent the final valid persuasiveness/   informativeness score. It is calculated as follows:

\begin{equation}
\begin{aligned}
\text {Per./Inf.}=\frac{1}{N} \sum_{i=1}^N\left(r_i \cdot GPTScore_i\right),\\
r_i= \begin{cases}1 & \text { if sample } i \text { refutes HS} \\ 0 & \text { otherwise }\end{cases},
\end{aligned}
\end{equation}
where $r_{i}$ denotes the detection result of $i$-th sample  by the CN classifier in SROC metric, $GPTScore_i$ represents the persuasiveness/informativeness of the $i$-th sample. $N$ is the number of generated CNs.    
    \item \textbf{Novelty} assesses the amount of novel content in generated
     CNs compared to the training corpus, calculated
following~\citet{wang2018sentigan}.

  \item \textbf{Linguistic quality}, as a basic requirement of text, is
   evaluated by the GRUEN metric~\citep{zhu2020gruen}, representing how
fluent, readable and grammatical the generated text is~\citep{zhu2021generate}.
 GRUEN utilizes a BERT-based model and a class of syntactic, semantic, and
contextual features to assesses the grammaticality, non-redundancy, focus and
structure/coherence of text. It is widely used to assess linguistic
quality~\citep{zhu2021generate,10.1145/3627106.3627122}.
\end{itemize}

\subsubsection{Human evaluation}
\textbf{Metrics.} Since human assessments can provide deeper understanding than automatic metrics, we randomly select 200 samples, each assigned to three proficient English annotators, to evaluate CNs from the following perspectives:
\begin{itemize}
  \item \textbf{Linguistic quality (LQ)} measures how fluent, readable, and grammatical the generated CN is.
  \item \textbf{Specificity (SPE)} refers to how directly or specifically the CN counters the central aspects of HS.
  \item \textbf{Choose-or-not (CON}) represents whether the generated CN can be
   used in real-case scenario~\citep{tekiroglu2022using}.
  \item \textbf{Is-best (BEST)} refers to whether the CN is the best one among
   the ones generated for the same HS~\citep{tekiroglu2022using}.
\end{itemize}

For LQ and SPE metrics, annotators are required to assign a score of 0 to
5~\citep{chung2020italian,tekiroglu2022using} to each sample. A higher score
\stmdocstextcolor{black}{represents} a better performance on the corresponding
metrics.
CHO and BEST indicators are binary ratings (1 or 0). For the CHO indicator, annotators need to mark the sample that they would use in the real scenario as 1 and the rest as 0. For the BEST indicator, the annotators need to mark the best generated CN for HS as 1 and the others as 0. The score $H$ for each metric is calculated as follows:
\begin{equation}
H=\frac{1}{M \times N} \sum_{i=1}^M \sum_{j=1}^N h_{ij},
\end{equation}
where $M$ represents the total number of annotators, $N$ represents the total number of annotated samples, and $h_{ij}$ denotes the score given by the $i$th annotator to the $j$th sample. 

\textbf{Quality Control.}
To ensure the quality of manual assessment, we first recruited annotators from
professional crowdsourcing agencies with a pass rate
\stmdocstextcolor{black}{of} over 90\% and proficiency in English.
Subsequently, based on~\citet{fanton2021human}, the selected annotators 
underwent the following training procedures, and three qualified annotators were chosen to perform manual assessments: 
\begin{enumerate}
    \item[(1)] Annotators should read and discuss NGO guidelines\footnote{\url{https://getthetrollsout.org/stoppinghate}.} 
    that describes CN writing campaigns against hate speech.
    \item[(2)]  Annotators should \stmdocstextcolor{black}{review} the expert-authored CN dataset to gain a better understanding of the attributes of CNs.
    \item[(3)]  Annotators are provided with explanations of the meaning and criteria for each evaluation metric, accompanied by bad and good instances for each metric to enhance their understanding of the metric's implications.
    \item[(4)] Before starting the formal annotation, annotators are asked to complete the qualification questionnaire in  \ref{exp:qq} of  \ref{sec:qq}.   
    Then, the annotators who passed the qualification questionnaire will be invited to conduct a round of \stmdocstextcolor{black}{tests} to label 10 instances and indicate the reasons. Any uncertainties arising during the test annotation will be discussed and resolved through meetings. Subsequently, the top three annotators whose annotations demonstrate a reasonableness rate exceeding 90\% and the most comprehensive justifications, are selected as formal annotators.

   \item[(5)] Throughout the annotation process, if annotators encounter problems, we will promptly convene meetings to facilitate discussions and resolutions.
\end{enumerate}

\section{Results and analysis}
\subsection{Automatic evaluation}
\stmdocstextcolor{black}{The experimental results of} the automatic metrics are shown in \ref{tab_exp:auto}. \stmdocstextcolor{black}{It can be observed that} ReZG surpasses both supervised and zero-shot baselines \stmdocstextcolor{black}{in} all metrics. It does not require \stmdocstextcolor{black}{an} in-target annotated corpora but can achieve better results than models that are trained on in-target CN \stmdocstextcolor{black}{corpora}. 
Compared to \stmdocstextcolor{black}{the} baselines, ReZG exhibits significant improvements in both \stmdocstextcolor{black}{the} relevance and success rate of countering, with \stmdocstextcolor{black}{increases of} 2.0\%+ and 4.5\%+ respectively. This indicates that ReZG \stmdocstextcolor{black}{can generate} CNs with heightened specificity.

\begin{table*}[pos=hbtp]%
\centering
\caption{Automatic evaluation results (in \%). The best results of baselines are underlined with wavy lines and those of all methods are shown in bold. ↑ /↓ means higher/lower scores are better for the corresponding metric. $\dagger$ (*) indicates that the improvements over baselines are statistically significant with $p < 0.001$ (0.05) for t-test. }
 \begin{tabular}{cccccccp{0.0001cm}cc}
    \hline
    \multicolumn{2}{c}{\multirow{2}{*}{\textbf{Method}}}
    &\multicolumn{5}{c}{\textbf{CN}}
    &
    &\multicolumn{2}{c}{\textbf{CN \& HS}} \\
    \cline{3-7}
    \cline{9-10}
\multicolumn{2}{c}{}
    & \textbf{Tox.↓ } & \textbf{Per.↑} & \textbf{Inf. ↑} & \textbf{Nov. ↑}  & \textbf{LQ ↑} &  &\textbf{Rel. ↑}& \textbf{SROC ↑}\\
   \hline
      \multirow{3}{*}{\begin{tabular}[c]{@{}c@{}}Supervised\\ Baselines\end{tabular}} & GPS  & \uwave{19.5} & 37.6& 25.7 &20.9 & 74.4 &  &4.8&73.6  \\
   \multirow{3}{*}{}    & GPT2-ft & 19.6 &\uwave{45.7}& \uwave{37.6} &\uwave{72.9} & \uwave{81.0}&  &14.6&83.9  \\
   \multirow{3}{*}{} & DialoGPT-ft & 21.2 & 43.1& 37.3 &70.8 &80.3& &13.8&86.4 \\
      \cline{1-10}
\multirow{8}{*}{\begin{tabular}[c]{@{}c@{}}Zero-shot\\ Baselines\end{tabular}}& T0-P$_{topk}$&22.1 & 43.2 & 28.6 & 64.5  &79.6 & &20.6 & 84.4 \\
  \multirow{8}{*}{} & T0-P$_{topp}$&20.9 & 38.7 & 25.4 & 63.0  &79.2 & &13.0 & 82.4 \\
  \multirow{8}{*}{} & T0-P$_{bs}$& 19.5 & 26.6 & 20.3 & 65.3  &79.0 & &18.9 & 58.2 \\
 \multirow{8}{*}{} &  T0-P$_{gs}$&27.9 & 32.3 & 26.0& 59.2 &\uwave{79.7} & &23.2  & 64.8  \\

\multirow{8}{*}{} &
  GPT2-P$_{topk}$&20.5 & 43.3 & 28.3 & 69.2 &79.1 && 13.3 & 82.7 \\

   \multirow{8}{*}{} & GPT2-P$_{topp}$&21.3 & 43.2 & 29.0& 70.0  &78.6 && 16.6  &86.6 \\
   \multirow{8}{*}{} &  GPT2-P$_{bs}$&\uwave{19.5} & 40.7 & 25.1   &64.8 &62.2 & &23.9& 81.3 \\
   \multirow{8}{*}{} & GPT2-P$_{gs}$&21.2 & 44.1 & 35.8& 71.8  &72.1 & & \uwave{26.0}& \uwave{87.5} \\
 \cline{1-10}
  \multirow{1}{*}{Ours}  &ReZG&\textbf{18.8$^{\dagger}$} & \textbf{47.4$^{\dagger}$} & \textbf{44.5$^{\dagger}$}& \textbf{73.5$^{\dagger}$}&\textbf{81.2$^{\dagger}$} & & \textbf{28.0$^{*}$} & \textbf{92.0$^{\dagger}$} \\
    \hline
  \end{tabular}
\label{tab_exp:auto}
\end{table*}

\stmdocstextcolor{black}{A comparison of} the supervised and zero-shot
baselines \stmdocstextcolor{black}{reveals} that the PLMs fine-tuned on the
in-target CN corpus generally \stmdocstextcolor{black}{have} higher effective
persuasion and information content, but a lower correlation with HS. This shows
to a certain extent that fine-tuning in the target domain can enable PLMs to
learn countering patterns for corresponding HS targets from the corpus.
\stmdocstextcolor{black}{Overall, the} average relevance and countering abilities
\stmdocstextcolor{black}{of T0-P for} HS are lower than those of GPT2-P. This
indicates that GPT2 \stmdocstextcolor{black}{has} stronger zero-shot ability in
understanding and perceiving the input than T0. This is reasonable
because\stmdocstextcolor{black}{,} without fine-tuning, the causal decoder LM
architecture usually outperforms the full encoder--decoder LM architecture
\stmdocstextcolor{black}{in} the zero-shot task~\citep{wang2022language}.
\stmdocstextcolor{black}{A} detailed analysis of \stmdocstextcolor{black}{the} samples generated by T0 \stmdocstextcolor{black}{shows that} T0-P$_{bs}$ and T0-P$_{gs}$ tend to produce duplicates or consistent content with HS.
They struggle to focus and fully comprehend the counter prompt \stmdocstextcolor{black}{effectively}.

Compared with T0 and GPT2, ReZG generates the least toxic content. We think this \stmdocstextcolor{black}{is} owing to the retrieved counter-knowledge, \stmdocstextcolor{black}{as} the counter-knowledge \stmdocstextcolor{black}{injected} into CNs is non-toxic and positive, which helps to guide \stmdocstextcolor{black}{the} PLM in generating non-toxic text.
Furthermore, by calculating the n-gram repetition rate and semantic similarity of the generated CNs and retrieved counter-knowledge, we find that generated CNs \stmdocstextcolor{black}{retain} 24.78\% of the retrieved counter-knowledge and \stmdocstextcolor{black}{achieve} a semantic similarity of 48.35\%.

The superior performance of ReZG in persuasiveness and informativeness \stmdocstextcolor{black}{is also} primarily attributed to the injected external counter-knowledge \stmdocstextcolor{black}{as} verified in the ablation study. This demonstrates the significance of counter-knowledge derived from external statistics, facts, or examples for CN generation. The results \stmdocstextcolor{black}{of} the relevance and SROC indicators demonstrate that ReZG not only \stmdocstextcolor{black}{can provide} a specific response to HS, but also \stmdocstextcolor{black}{has} a strong ability to counter it.

\subsection{Human evaluation}
\begin{table*}[pos=htbp]%
\centering
\caption{Human evaluation results of ReZG and baselines. The best results are shown in bold. ↑ means higher scores are better for the corresponding metric.}
 \begin{tabular}{cccccc}
    \hline
 \multicolumn{2}{c}{\textbf{Methods}}
  & \multicolumn{1}{c}{\textbf{LQ ↑}}
   &\multicolumn{1}{c}{\textbf{SPE ↑}}
    &\multicolumn{1}{c}{\textbf{CON ↑}}
    &\multicolumn{1}{c}{\textbf{BEST ↑}}
    \\
    \hline                                       
  \multirow{3}{*}{\begin{tabular}[c]{@{}c@{}}Supervised\\ Baselines \end{tabular}} &    
GPS &3.801 &1.013 &0.470 & 0.093\\
\multirow{3}{*}{} &   
GPT2-ft  &4.259  &2.264 &0.463 & 0.100 \\
\multirow{3}{*}{} &   
DialoGPT-ft  &4.248 &2.257 & 0.480 & 0.112 \\
\cline{1-6} 
  \multirow{8}{*}{\begin{tabular}[c]{@{}c@{}}Zero-shot\\ Baselines \end{tabular}} &  
  T0-P$_{topk}$&4.017 & 2.962&0.468 &0.110 \\
\multirow{8}{*}{} &  
T0-P$_{topp}$&4.013 & 2.760 & 0.322&0.072  \\
\multirow{8}{*}{} &   
T0-P$_{bs}$& 3.665 &1.932 &0.260 &0.067 \\
\multirow{8}{*}{} &   
T0-P$_{gs}$& 3.761 &2.383 &0.365 & 0.058\\
 
\multirow{8}{*}{} &   
 GPT2-P$_{topk}$&3.980 & 2.677 & 0.231 & 0.042 \\
  \multirow{8}{*}{} &  
GPT2-P$_{topp}$&4.247 & 2.822 & 0.298   & 0.063 \\
  \multirow{8}{*}{} &    
  GPT2-P$_{bs}$&4.015  &2.775 &0.472 & 0.085 \\
  \multirow{8}{*}{} &    
  GPT2-P$_{gs}$& 4.182 &3.137 &0.412 & 0.075\\
\cline{1-6}    
  \multirow{1}{*}{Ours} &   
ReZG & \textbf{4.345} & \textbf{3.320} &\textbf{0.482}& \textbf{0.123}  \\
 \hline
 \multicolumn{2}{c}{\textcolor{black}{Fleiss' Kappa:  $k$}} &0.658 & 0.542 & 0.825 &0.831\\
    \hline
    
  \end{tabular}
\label{tab_exp:human}
\end{table*} 
We present the results of human evaluation in 
Table \ref{tab_exp:human}. Consistent with the results of automatic evaluation, the CNs generated by ReZG have a higher specificity and can counter the central part of the HS more accurately while maintaining good fluency, compared to the ones generated by baseline. The superior CON and BEST scores for ReZG indicate that CNs generated by ReZG are most likely to be used by researchers in real scenarios. This also shows from the side that CNs generated by ReZG are more convincing, and verifies the effectiveness of GPT-3.5-turbo in evaluating persuadability and other indicators.
In terms of inter-annotator agreement, the annotators show almost perfect agreement on CON and BEST metrics \stmdocstextcolor{black}{($k \in [0.81,1.00]$)}, and substantial agreement on the linguistic quality metric (\stmdocstextcolor{black}{($k \in [0.61,0.800]$)}) and \stmdocstextcolor{black}{moderate agreement on ($k \in [0.41,0.60]$)} the specificity metric. \stmdocstextcolor{black}{Some examples of the human evaluation results are illustrated in the 
Tables \ref{tab:human_example1}, \ref{tab:human_example2}  and \ref{tab:human-example3}. of Appendix. \ref{sec:humam_example}.}

\subsection{Analysis on generalization capabilities}
 \begin{figure}[thbp]
  \centering
  \includegraphics[width=.45\textwidth]{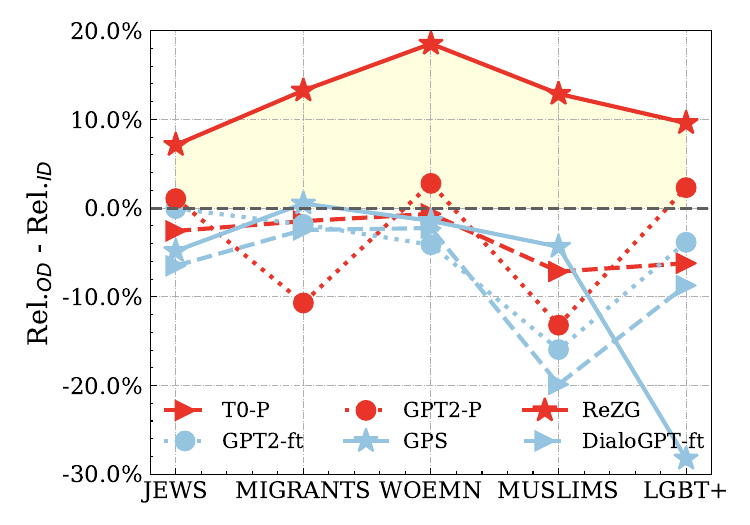}
  \includegraphics[width=.45\textwidth]{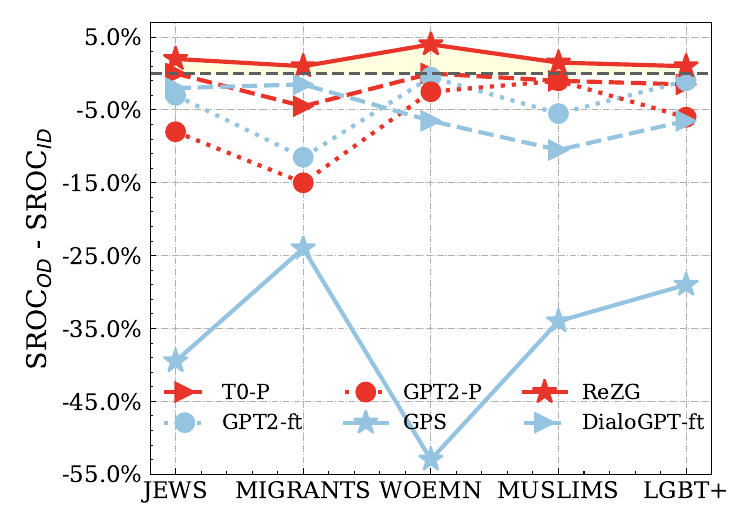}
  \caption{Relative SROC and relevance scores between in-domain and out-of-domain models. The abscissa represents the test domain of each target. Rel$_{OD}$ and Rel$_{ID}$ represent the relevance score of out-of-domain and in-domain models, respectively. SROC$_{OD}$ and SROC$_{ID}$ denote the SROC score of the in-domain and out-of-domain models, respectively.}
  \label{exp:ood}
\end{figure}
To analyze the generalization capabilities of the ReZG, we build a set of in-domain and out-of-domain experiments on the ReZG and baselines. Specifically, we focus on the relevance and SROC performance of the models on different unseen targets because the generated samples are required to capture and counter the central aspects of the corresponding HS.  
For the GPT2-P and T0-P baselines, we select the greedy search and top-$k$ decoding mechanisms separately because they have the best performance on the SROC metric among all decoding algorithms.

First, the targets with the highest number of samples are selected as test domains (\verb"MIGRANTS," \verb"JEWS," \verb"WOMEN," \verb"LGBT+," and \verb"MUSLIMS") to ensure a sufficient sized test set.
Then, we randomly select 200 samples from each test domain as the corresponding test cases. In-domain and out-of-domain models share the same test set. For in-domain models, we compose training sets by randomly sampling 400 samples from each test domain, resulting in a total of 2,000 samples. For out-of-domain models, we remove samples of the respective target from the aforementioned training set for each test domain. Subsequently, to balance the experimental setting, we randomly select 400 samples from the \verb"Disabled" and \verb"POC" domains to fill into the training set.
Because no training process is available for zero-shot baselines, the best outcomes of the in-domain models are used as benchmarks for comparison.

Fig. \ref{exp:ood} illustrates the difference in relevance and SROC scores between the in-domain and out-of-domain models for each method.
It reveals that ReZG outperforms both the supervised and zero-shot baselines in terms of generalizability, whereas GPS shows the lowest generalizability.
We believe that the strong generalizability of ReZG is most likely attributed to the reason that retrieval augmentation and a powerful PLM as a base model make it less susceptible to out-of-distribution influences.

\subsection{Analysis \textcolor{black}{of} SSF algorithm}
\label{sec:ssf}
\begin{figure}[htbp]
  \centering
  \includegraphics[width=.5 \textwidth]{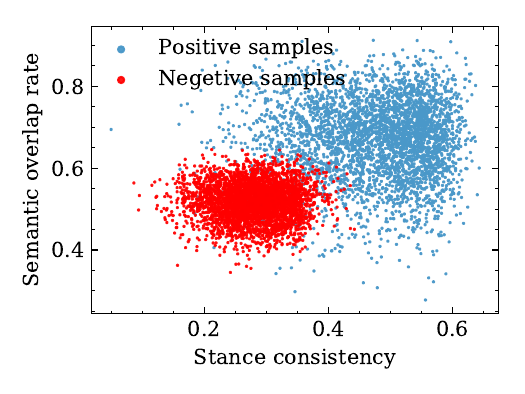}
  \caption{Distribution of positive and negative samples in stance consistency and semantic overlap rate.}
  \label{fig:sta-sem}
\end{figure}

\paragraph{\textbf{STA-SEM \textcolor{black}{d}istribution of samples.}}
\stmdocstextcolor{black}{To} verify whether the SSF algorithm can distinguish invalid and valid information, we \stmdocstextcolor{black}{calculate} the semantic overlap rate and stance similarity between \stmdocstextcolor{black}{the} positive and negative samples on the expert-authored CN dataset, as shown in  \ref{fig:sta-sem}. For the semantic overlap rate, positive samples refer to CNs for the same HS, and CNs of different HS are negative samples to each other. Given that the stance consistency is for antecedent statements, HS of the same target is regarded as a positive sample and HS with different targets is regarded as a negative sample. 
 \ref{fig:sta-sem} clearly illustrates that a notable distinction in the distribution between the positive and negative samples. The positive samples exhibit higher levels of semantic overlap and stance consistency compared \stmdocstextcolor{black}{with} \stmdocstextcolor{black}{the} negative samples, thereby verifying the rationality of retrieving useful information based on \stmdocstextcolor{black}{the} SEM and STA criteria within the SSF algorithm.

\begin{figure*}[htbp]
  \centering
  \subfigure[\textcolor{black}{The relevance score between CN and HS.}]{
  \includegraphics[width=.4 \textwidth]{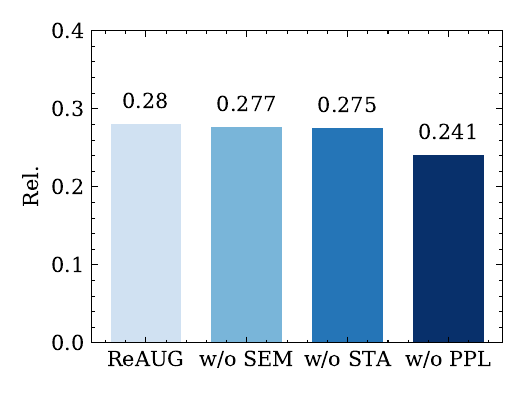}}
  \subfigure[The countering success rate of CN to HS.]{
    \includegraphics[width=.4 \textwidth]{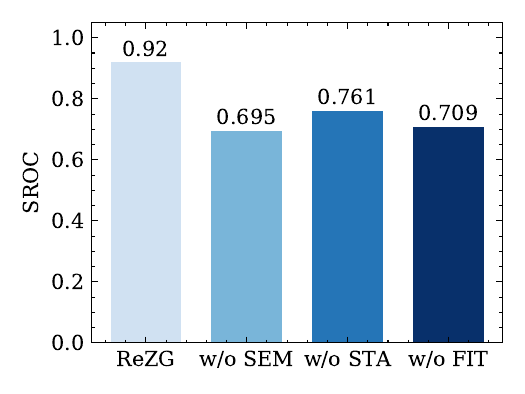}}
  \caption{The experimental results of respectively ablating semantic overlap rate, stance consistency and fitness in SSF retrieval algorithm.}
  \label{fig:ssf-abltion}
\end{figure*}

\paragraph{\textbf{Dimension ablation of SSF algorithm.}}
\stmdocstextcolor{black}{To} explore the effectiveness of the features of each dimension in the SSF algorithm, we remove the semantic overlap rate, stance consistency and fitness features in the SSF algorithm from ReZG \stmdocstextcolor{black}{respectively}, denoted as w/o SEM, w/o STA and w/o FIT. 
As shown in Fig. \ref{fig:ssf-abltion},  the absence of any dimension \stmdocstextcolor{black}{weakens} the final generative performance, validating the effectiveness of each dimensional feature. When the fine-grained filtering based on fitness is removed, the relevance between the generated CN and HS decreases the most. This is because the content filtered by SEM and STA contains more invalid and redundant information, which \stmdocstextcolor{black}{distracts the model's attention}. FIT screens out more accurate external knowledge from the perspective of language modeling based on perplexity. This \stmdocstextcolor{black}{further} demonstrates the necessity of fine-grained retrieval based on fitness.

\paragraph{\textbf{SSF vs. other retrieval methods.}} 
In this section, we compare the SSF algorithm with
the widely-used sparse retriever BM25~\citep{robertson2009probabilistic} and
dense retriever DPR~\citep{karpukhin2020dense}, to analyze its enhancements
\stmdocstextcolor{black}{for} the CN generation task. 
Specifically, we first \stmdocstextcolor{black}{conduct experiments by replacing} the CKR modules in ReZG with \stmdocstextcolor{black}{the} BM25 and DPR retrievers.
\stmdocstextcolor{black}{Subsequently,} to ascertain whether similar outcomes
can be achieved with generators other than \stmdocstextcolor{black}{the} EnG,
we replace the EnG with Llama2-70B-chat~\citep{touvron2023Llama} and use the
prompt learning to instruct Llama2-70B-chat to use the retrieved information as a
context to generate CNs. \stmdocstextcolor{black}{The simple prompt, SimPrompt
in Table \ref{tab_exp:eng_prompt}, is used as the input instruction,
considering that the base model of EnG, GPT2, has not undergone instruction
tuning and struggles to handle complex instructions.
}

As shown in the experimental results in Table \ref{tab:ssf-ablation}, the SSF retriever exhibits superior scores across both generators, indicating its effectiveness in retrieving counter-knowledge compared to traditional retrievers.
\stmdocstextcolor{black}{A comparison of } the results of Llama2-70B-chat and the proposed EnG under the same retriever \stmdocstextcolor{black}{shows} that \stmdocstextcolor{black}{the} EnG achieves better performance in \stmdocstextcolor{black}{the} relevance and SROC than the Llama2-70B-chat generator, whose parameters are dozens of times larger. This is reasonable because such PLMs \stmdocstextcolor{black}{do not perform} explicit training on integrating external knowledge. This demonstrates that the proposed EnG excels in effectively integrating retrieved external information.

\begin{table}[hbtp]
\centering
\caption{Comparative experimental results (in \%) between SSF retriever and other retrievers.}
\begin{tabular}{c|c|cc}
\hline
\multicolumn{1}{c|}{\multirow{2}{*}{\textbf{Retrieval Method}}} & \multirow{2}{*}{\textbf{CN generator}} & \multicolumn{2}{c}{\textbf{CN\&HS}} \\ \cline{3-4} 
\multicolumn{1}{c|}{} &  & \textbf{Rel.} & \textbf{SROC} \\ \hline
BM25 & \multirow{3}{*}{\begin{tabular}[c]{@{}c@{}}+ EnG \\ generator\end{tabular}} & 27.9 & 56.1 \\
DPR &  & 22.1 & 46.6 \\
SSF &  & \textbf{28.0} & \textbf{92.0} \\ \hline
BM25 & \multirow{3}{*}{\begin{tabular}[c]{@{}c@{}}+ Llama2-70B-chat \\  generator\end{tabular}} & 7.1 & 65.3 \\
DPR &  & 8.8 & 64.8 \\
SSF &  & \textbf{15.8} & \textbf{71.9} \\ \hline
\end{tabular}
\label{tab:ssf-ablation}
\end{table}

\subsection{Analysis \textcolor{black}{of} EnG module}
\label{sec:eng_ex}
\paragraph{\textbf{Ablation analysis of constraint functions.}}In to further investigate the impact of every constraint function in EnG, we conduct comparison between ReZG and its variants, namely \textbf{ReZG$_{f_{\operatorname{flu}}}$}, \textbf{ReZG$_{f_{\operatorname{cc}}}$}, \textbf{ReZG$_{f_{\operatorname{sim}}}$}, \textbf{ReZG$_{f_{\operatorname{flu}}+f_{\operatorname{cc}}}$}, \textbf{ReZG$_{f_{\operatorname{cc}}+f_{\operatorname{sim}}}$} and \textbf{ReZG$_{f_{\operatorname{flu}}+f_{\operatorname{sim}}}$}.  
The former three variants generate CNs only using the fluency constraint, countering constraint, or knowledge preservation constraints respectively. The latter three variants generate CNs with corresponding two constraints.

 \begin{figure*}[htbp]
    \centering
   \includegraphics[width=0.4\textwidth]{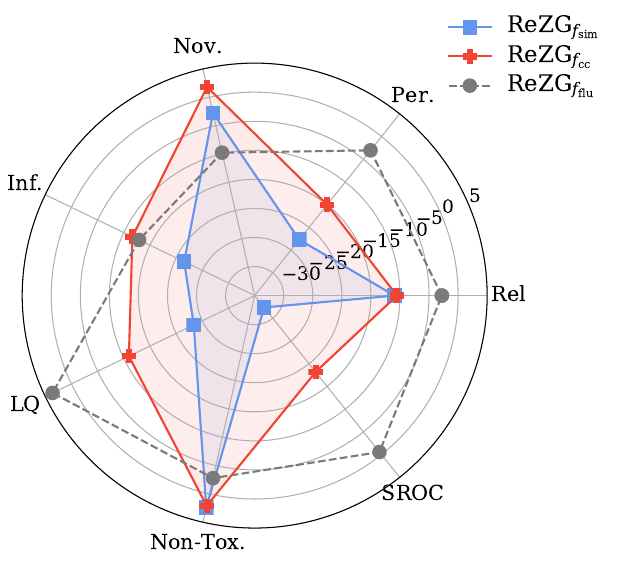}
    \includegraphics[width=0.42\textwidth]{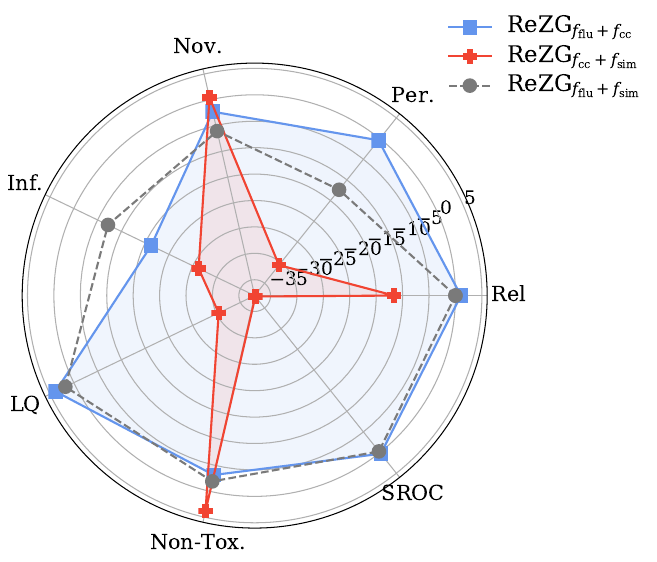}  
    \caption{Ablation results of constraint functions. It represents score differences (in \%)  between ablations and ReZG on seven automatic evaluation metrics. NoT-Tox. equals the negative of the toxic score difference.}
    \label{fig:constraint}
\end{figure*}

The results of subtracting \textcolor{black}{the} scores of ablations from \textcolor{black}{the} scores of ReZG \textcolor{black}{in the} automatic evaluation metrics are shown in Fig.~\ref{fig:constraint}. From this figure, we can intuitively \textcolor{black}{observe} the impact of each constraint on the \textcolor{black}{metrics} of the generated CN\textcolor{black}{s}. The right subgraph of Fig.~\ref{fig:constraint}  illustrates that reducing any constraint will lead to a decline in the persuasiveness, informativeness, relevance\textcolor{black}{,} and countering success rate of the generated CNs.

\textbf{Fluency constraint.} \textcolor{black}{The} outcomes of ReZG$_{f_{\operatorname{cc}}+f_{\operatorname{sim}}}$ demonstrate that upon the removal of \textcolor{black}{the} fluency \textcolor{black}{constraint}, the performance of \textcolor{black}{the} generated CNs declines \textcolor{black}{markedly} across all metrics excluding novelty and non-toxicity. This is mainly because\textcolor{black}{, when} the fluency constraint is removed, the generated samples become unreadable and messy, resulting in a significant decline in \textcolor{black}{the} language quality and SROC. \textcolor{black}{In addition} the decline in \textcolor{black}{the} SROC precipitates reductions in persuasiveness and informativeness. Similar results \textcolor{black}{can be observed for} on ReZG$_{f_{\operatorname{cc}}}$ and Re ZG$_{f_{\operatorname{sim}}}$ which \textcolor{black}{do not include the} fluency constraint. This demonstrates the indispensability of the fluency constraint as a fundamental requirement for CN generation, despite their potential trade-offs with \textcolor{black}{the} novelty and non-toxicity metrics.

\textbf{Knowledge preservation constraint.} For the variant ReZG$_{f_{\operatorname{flu}}+f_{\operatorname{cc}}}$ that does not \stmdocstextcolor{black}{include} the knowledge preservation constraint, it exhibits a significant reduction \stmdocstextcolor{black}{in} effective informativeness even when its SROC remains relatively high. Conversely,  ReZG$_{f_{\operatorname{flu}}+f_{\operatorname{sim}}}$ \stmdocstextcolor{black}{which includes the} knowledge preservation constraints\stmdocstextcolor{black}{, exhibits a} minimal decrease in effective informativeness with a similar SROC score. This proves the effectiveness of the knowledge preservation constraint \stmdocstextcolor{black}{in fusing} external information and the SSF algorithm \stmdocstextcolor{black}{in acquiring} counter-knowledge. 
Meanwhile, the results of ReZG$_{f_{\operatorname{sim}}}$ demonstrate that the knowledge preservation constraint also enhances the novelty and non-toxicity of CNs.
It is logical \stmdocstextcolor{black}{that} the lowest SROC \stmdocstextcolor{black}{is observed} with only the knowledge preservation constraint, as the absence of a fluency constraint often leads to incoherent CNs that retain counter-knowledge haphazardly, causing the classifier to label them as unsuccessful cases. This also proves that a single constraint is inadequate for achieving optimal outcomes, and a combination of multiple constraints is required.

\textbf{Countering constraint.} The results of ReZG$_{f_{\operatorname{flu}}+f_{\operatorname{sim}}}$ 
show that\stmdocstextcolor{black}{,} when the countering constraint is removed, \stmdocstextcolor{black}{the} persuasiveness \stmdocstextcolor{black}{decreases markedly} even when the SROC is high.
This suggests that the countering constraint enhances the persuasiveness of generated CNs, playing\stmdocstextcolor{black}{s an} important role in guiding the generation of target sequences towards countering HS.

By comparing the results of ReZG with its variants under single and dual constraints, it is \stmdocstextcolor{black}{observed} that the integration of multiple constraints can enhance the model performance from many aspects, leading to a more comprehensive improvement in the attributes of \stmdocstextcolor{black}{the} generated CNs.

\paragraph{\textbf{EnG vs. other generators.}}
To determine \stmdocstextcolor{black}{the} effectiveness of
\stmdocstextcolor{black}{the} EnG, \stmdocstextcolor{black}{we replace the EnG
in ReZG with the base model GPT2~\citep{radford2019language} and five advanced
LLMs with powerful zero-shot generation capabilities for comparison. These five
LLMs are  GPT-3.5-Turbo~\cite{gpt3}, GPT-4o~\cite{openai2024gpt4o},
Qwen2-72B-Instruct~\cite{yang2024qwen2}, Llama2-70B-chat
~\citep{touvron2023Llama} and Llama3.1-70B-Instruct~\cite{dubey2024Llama}.}
\stmdocstextcolor{black}{In addition, we employ three different prompts, which are the SimPrompt, EnGPrompt, and KnoPrompt in 
Table \ref{tab_exp:eng_prompt} of  \ref{sec:eng_prompt}, to instruct the LLMs to generate CNs with the retrieved information.
The SimPrompt is the simple prompt used in  \ref{sec:ssf}, which guides the LMs to consider the retrieved information to generate CNs.
The EnGPrompt incorporates the same constraints as EnG's restrictions. 
The KnoPrompt is the knowledgeable dialog generation prompt proposed 
in the Multi-Stage Dialog Prompting (MSDP) framework~\cite{liu2022multi},
which excels at generating knowledge-rich responses based on the dialog
context and the retrieved knowledge.}

\begin{table*}[pos=htbp]
\centering
\caption{\textcolor{black}{Comparative experimental results (in \%) between EnG and other generators. The best results of baselines are underlined with wavy lines and those of all methods are shown in bold.}}
\setlength{\tabcolsep}{3pt}
\begin{tabular}{c|c|ccc|ccc}
\hline
\multirow{2}{*}{\textbf{\begin{tabular}[c]{@{}c@{}}Retrieval \\ Method\end{tabular}}} & \multirow{2}{*}{\textbf{CN generator}} & \multicolumn{3}{c|}{\textbf{Rel.}}                                          & \multicolumn{3}{c}{\textbf{SROC}}                                         \\ \cline{3-8} &                                        & \textbf{SimPrompt} & \textbf{EnGPrompt}        & \textbf{KnoPrompt}         & \textbf{SimPrompt}        & \textbf{EnGPrompt}        & \textbf{KnoPrompt} \\ \hline
\multirow{7}{*}{SSF}  & GPT2  & 8.2 & 6.5       & 0.0   & 52.3   & 58.0    & 12.8      \\
& GPT-3.5-turbo        & 11.4   & 27.5               & 21.7   & 67.3      & 89.5               & 68.8               \\
& GPT-4o   & 21.0   & 24.3  & 21.7   & 87.0 & 89.8   & 85.2     \\
& Qwen2-72B-Instruct     & 16.2   & 14.6               & \uwave{24.9}     & 84.2     & 90.3    & 78.4             \\
& Llama2-70B-chat  & 15.8   & 24.4    & 18.5               & 71.9   & 83.2  & 55.7  \\
& Llama3.1-70B-Instruct  & 18.9      & 21.0               & 10.4   & 77.3   & \uwave{90.9}               & 67.0               \\ \cline{2-8} 
& EnG    & \multicolumn{3}{c|}{\textbf{28.0}}    & \multicolumn{3}{c}{\textbf{92.0}}                            \\ \hline
\end{tabular}
\label{tab:generator-abltaion}
\end{table*}

As shown \textcolor{black}{in the experimental results} in Table \ref{tab:generator-abltaion}, the relevance and SROC scores \textcolor{black}{exhibit a} substantially decrease after substituting the EnG with \textcolor{black}{the} baseline generators. This demonstrates that the proposed EnG excels in effectively integrating counter-knowledge in retrieved external information, justifying \textcolor{black}{its} superior effectiveness.
\stmdocstextcolor{black}{The baselines achieve the highest SROC score of 90.9\% under the EnGPrompt with the Llama3.1-70B-Instruct model, while the Qwen2-72B-Instruct model attains the highest relevance score of 24.9 under the KnoPrompt. Notably, the performance of the proposed EnG generator exceeds the best results of all baselines, achieving improvements of 3.1\%+ in the relevance score and 1.1\%+ in the SROC score.}

\stmdocstextcolor{black}{Comparing the results of different prompts on the same generator, we can find that the EngPrompt achieves the highest SROC in all generators and the highest relevance score in all generators except for the Qwen2-72B-Instruct. This is reasonable because its instruction contains the constraint for retaining counter-knowledge, which is essential for CN generation, while other prompts lack such constraint. This further supports the rationality of the constraint settings in the EnG generator.
In contrasting the results of different generators under the same prompt, it is apparent that the relevance and SROC of the CNs generated by GPT2 are significantly lower than those of other generators under all prompts. This disparity arises because the parameter count of GPT2 is substantially smaller than that of other generators and it has not undergone instruction tuning, which results in limited knowledge retention and reasoning capabilities. Consequently, it struggles to comprehend and execute complex task instructions effectively. EnG, by incorporating multiple constraints based on the energy function into GPT2, enables it to achieve performance beyond these powerful LLM baselines on the retrieval-augmented  CN generation task.}

\subsection{\textcolor{black}{Impact analysis of CMV knowledge repository}}
\stmdocstextcolor{black}{To investigate the impact of external information <$\mathbb{P},\mathbb{C}$> from CMV on model performance, we randomly sample the same number of instances from <$\mathbb{P},\mathbb{C}$> as the original <$HS,CN$> training dataset to create a new training dataset composed only of CMV data. This newly constructed dataset, with uniformly distributed samples across all target classes, is denoted as CMV-T. The original <$HS,CN$> training set is denoted as BASE. We then conduct comparative experiments by using CMV-T, BASE, and the union of CMV-T and BASE respectively as the training datasets for the supervised baseline models.}
  
\stmdocstextcolor{black}{The ablation results of baselines under different training sets are shown in 
Table \ref{tab:cmv-t}. It can be seen that the models trained solely on the CMV data achieve a much lower SORC compared to the models trained on the expert-curated CN dataset. When the training dataset includes both BASE and CMV-T sets, the model's SROC metric decreases slightly by less than 1.0\%, while the relevance scores increase remarkably by approximately 1\% to 15\%, compared to using the BASE dataset alone. This suggests that CMV-T data can enhance the relevance between generated CNs and HS, but do not contribute to improving the countering success rate. This is most likely because the most counter-comments in CMV-T closely narrate around the input posts, thus exhibiting a strong relevance, which affects the pattern of CN generation. Meanwhile, the ReZG makes selective and controlled use of CMV data through the SSF retrieval method and constrained decoding mechanism, enabling it to significantly outperform the baseline models trained on CMV data.}

\begin{table}[pos=htbp]
\centering
\caption{\textcolor{black}{Ablation results (in \%) on the training dataset.}}
\begin{tabular}{c|c|cc}
\hline
\multirow{2}{*}{\textbf{Training Set}} & \multirow{2}{*}{\textbf{method}} & \multicolumn{2}{c}{\textbf{CN\&HS}} \\ \cline{3-4}  &  & \textbf{Rel.}    & \textbf{SROC}    \\ \hline
\multirow{3}{*}{CMV-T}   & GPS    &  15.5   & 61.1   \\
         & GPT2-ft      & 18.6      & 82.7       \\
      & DialoGPT-ft    & 11.1             & 75.0         \\ \hline
\multirow{3}{*}{BASE}                  & GPS                              & 4.8              & 73.6             \\
        & GPT2-ft         & 14.6    & 83.9             \\
      & DialoGPT-ft         & 13.8   & 86.4             \\ \hline
\multirow{3}{*}{CMV-T\&BASE}      & GPS                              &     16.0             &     73.3             \\
        & GPT2-ft    & 15.5   & 83.0     \\
        & DialoGPT-ft   & 15.8       & 85.8    \\ \hline
-    & ReZG          & 28.0      & 92.0   \\ \hline
\end{tabular}
\label{tab:cmv-t}
\end{table}

\subsection{Performance of ReZG on different types of HS}
\begin{figure*}[th!]
  \centering
      \includegraphics[width=.3\textwidth]{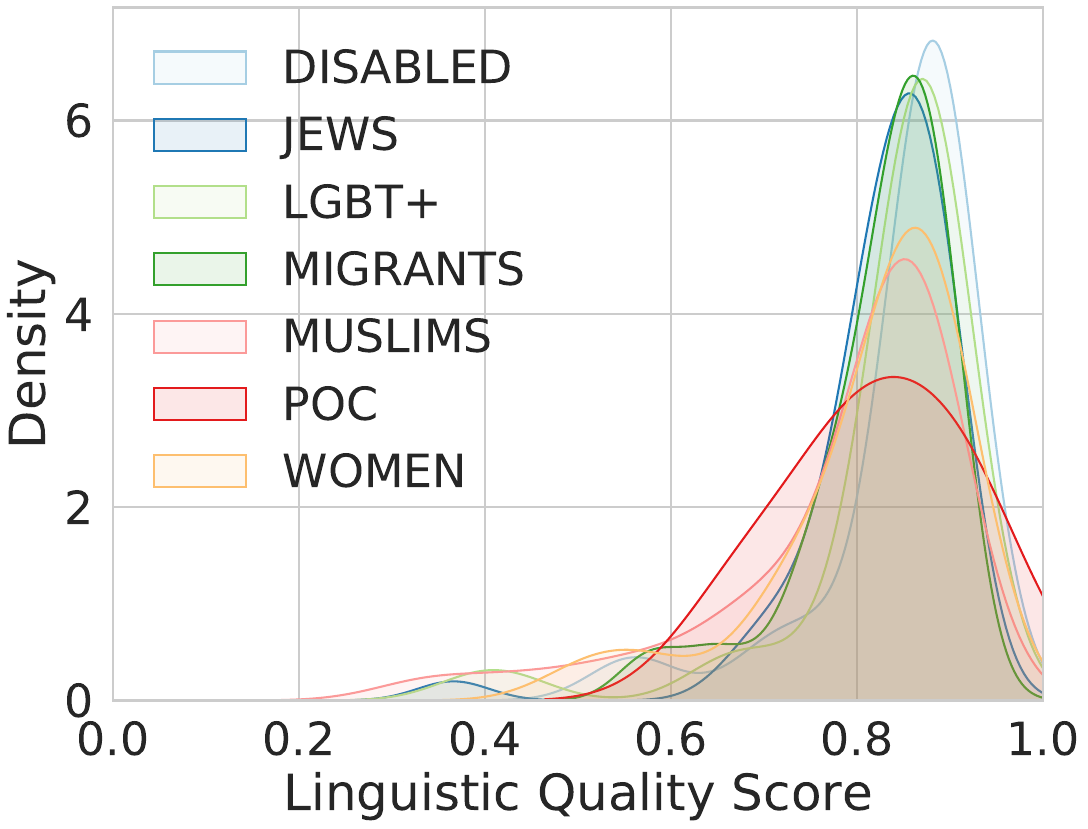}   
      
    \includegraphics[width=.3\textwidth]{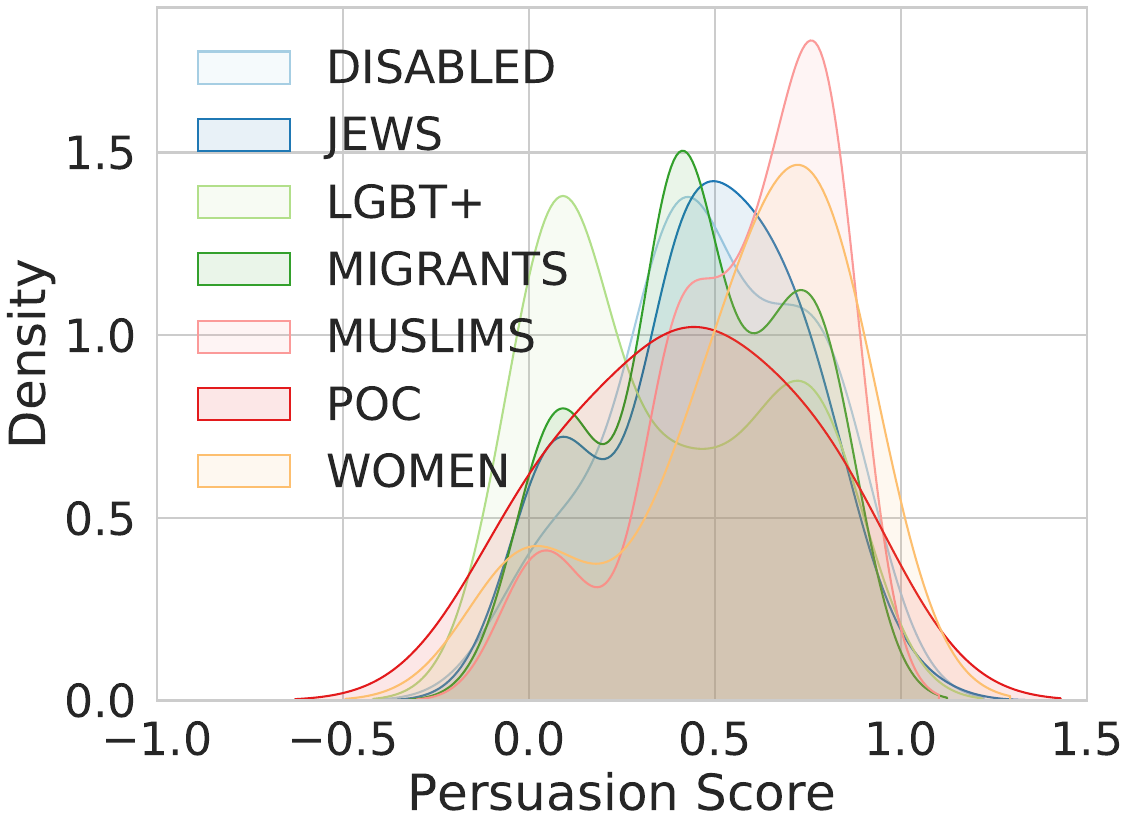}
    \includegraphics[width=.31\textwidth]{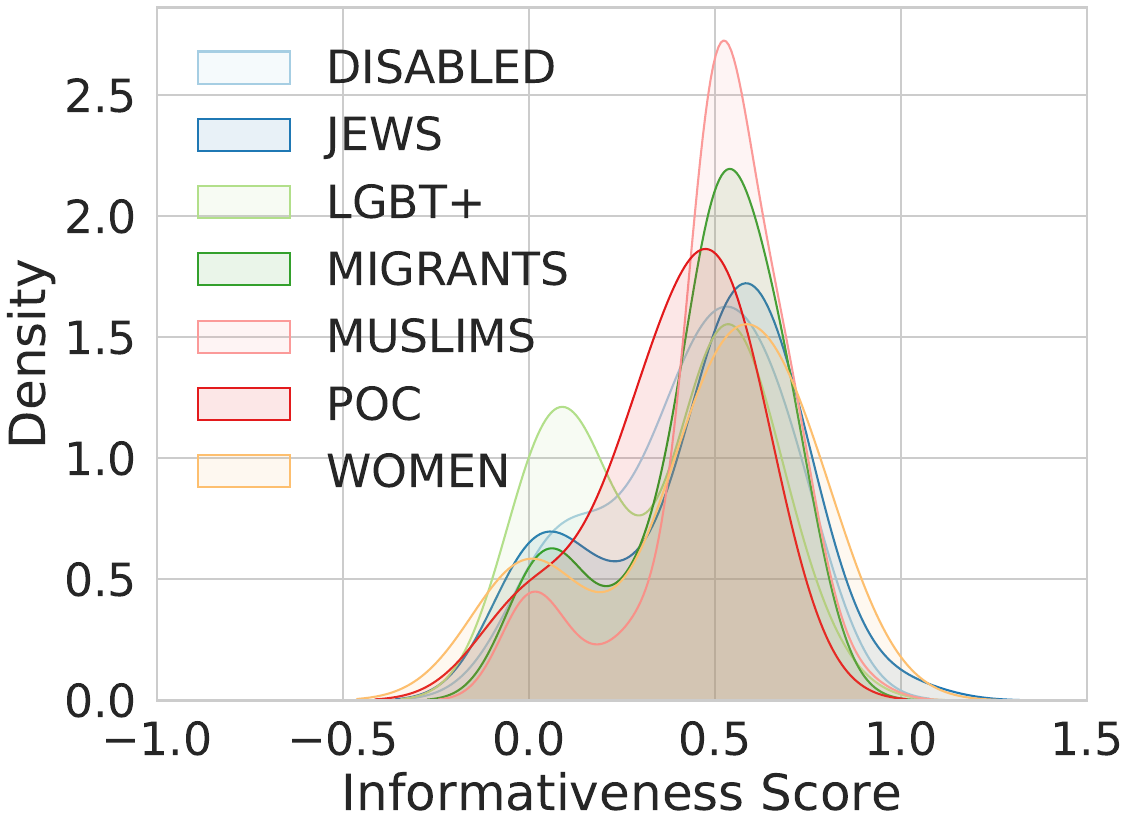}
    \includegraphics[width=.3\textwidth]{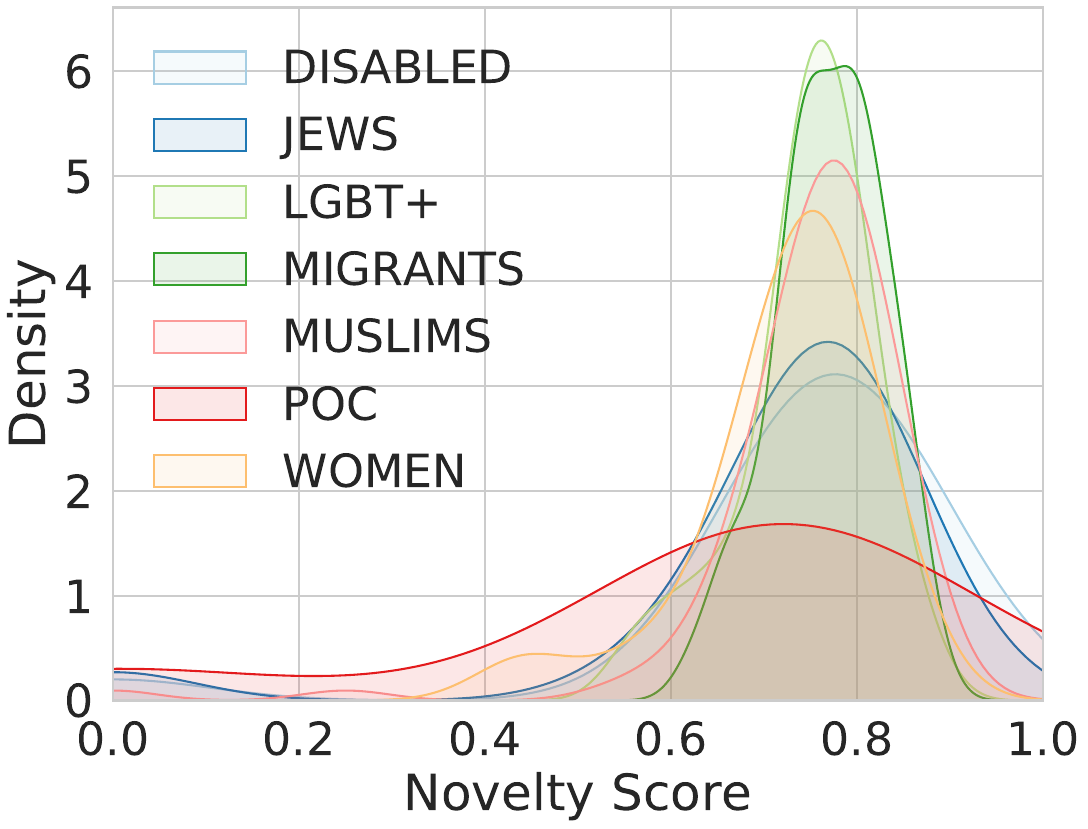}
    \includegraphics[width=.3\textwidth]{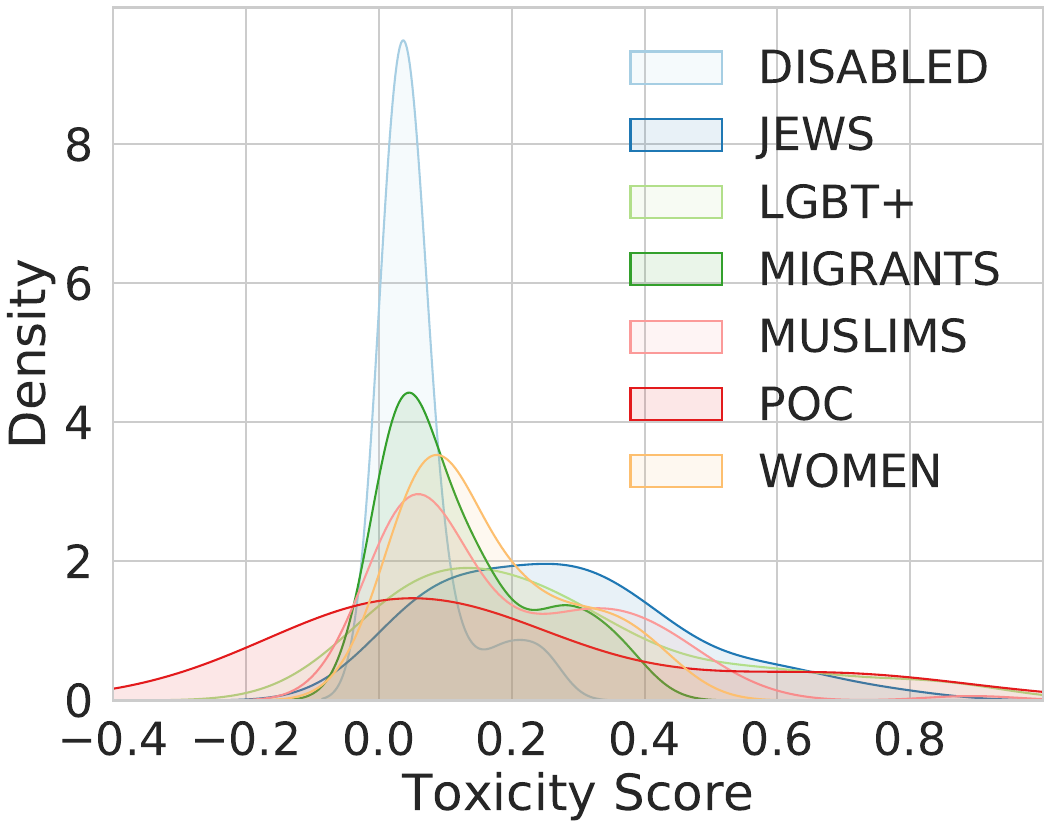}
  \includegraphics[width=.31\textwidth]{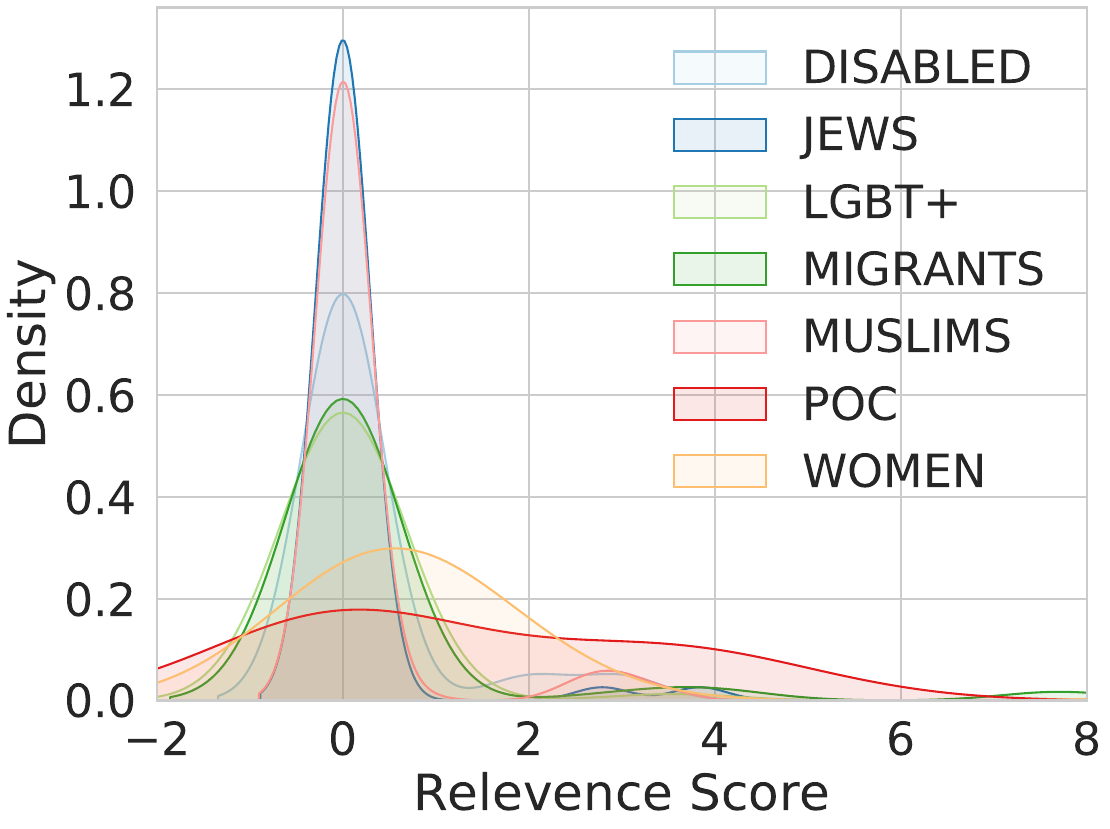}
    \includegraphics[width=.3\textwidth]{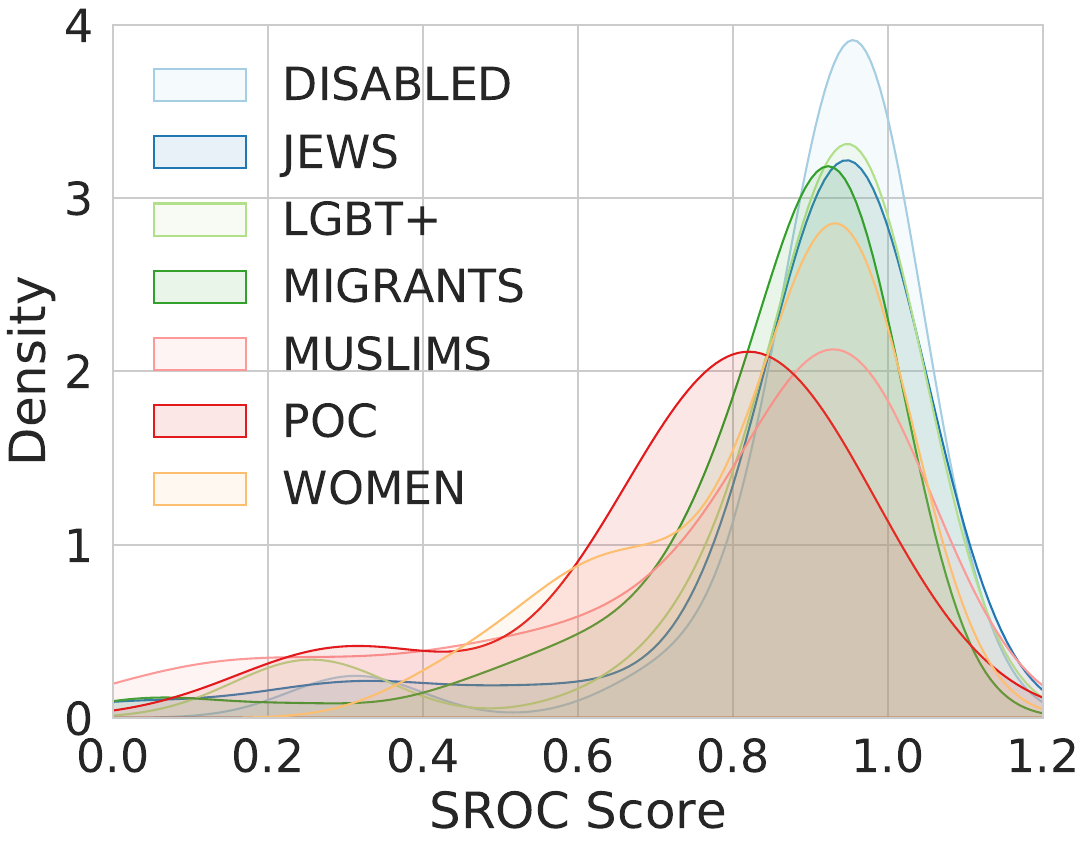}

  \caption{Score density distributions of ReZG with different types of HS.}
  \label{fig:ReZG_types}
\end{figure*}
To further analyze the performance of ReZG in terms of different HS targets, we calculate the density distributions of evaluation results for linguistic quality, toxicity, persuasion, informativeness, etc., as shown in Fig. \ref{fig:ReZG_types}.

The results demonstrate that ReZG achieves a better and more concentrated  density distribution in all metrics when generating CNs for HS targeting the Disabled. This suggests that ReZG is more effective at countering HS targeted towards the disabled. Meanwhile, the density distribution of the POC exhibits the highest dispersion across all metrics, primarily due to its dispersion among groups with different skin colors. 
Compared to other HS targets, the generated CNs of POCs exhibit a lower linguistic quality, novelty, and the success rate of countering. Addressing HS of POC target poses heightened complexity and challenges.

\subsection{Case study}
In this section, we present a qualitative analysis of \stmdocstextcolor{black}{a} case study to clearly interpret the performance of ReZG. Examples of CNs generated by ReZG and baselines are presented in 
Table \ref{tab_exp:case}.

As shown in 
Table \ref{tab_exp:case}, the CN generated by ReZG is the closest to that written by experts. It not only rejects the view expressed by the HS, but also powerfully expounds the reasons for disagreement. We also find that the samples generated by GPT2-P$_{bs}$, GPT2-P$_{gs}$, and T0-P$_{gs}$ all fail to counter the HS, indicating that it is difficult to ensure that the generated samples can counter HS expressions by relying solely on the counter prompt. Meanwhile, the sample generated by GPT2-ft reveals that even supervised methods may produce toxic content, \stmdocstextcolor{black}{since the model is inevitably exposed to harmful HS-related language during pre-training and fine-tuning}. In contrast, the CN generated by ReZG is more specific, persuasive, and closer to that written by experts.

\begin{table*}[b!]
\centering
\caption{Examples of generated CNs against hate speech. Among them, the \textcolor{violet}{violet examples} have explained the reasons for countering HS, the \textcolor{Green}{green examples} hold the same view with HS, and the \textit{italic example} represents containing toxic content.}
 \begin{tabular}{l|p{11cm}}
    \hline                                        
    \textbf{Hate Specch} &Shaming Fatties is just for their good. \\
    \hline
     \textbf{Expert-authored} &Even if not all shaming carries negative meaning, \textcolor{violet}{it is difficult for individuals to distinguish between shaming that comes from a place of concern as opposed to discrimination}. \\
     \hline
    \textbf{GPT2-P$_{topk}$} &I don't want to shame the obese, I don't want to shame the fat people. \\
    \textbf{GPT2-P$_{topp}$}  & I don't like shaming anyone. It's not my place to make that decision for anyone. \\
    \textbf{GPT2-P$_{bs}$}  &\textcolor{ForestGreen}{I think shaming people for being fat is just for their good.} \\
    \textbf{GPT2-P$_{gs}$} &\textcolor{ForestGreen}{I think shaming is a very effective way to get people to change their behavior.} \\
    \textbf{T0-P$_{topk}$}  & I don't believe fat people should be shaming. \\
    \textbf{T0-P$_{topp}$}  & It's degrading to the overweight and obese. \\
    \textbf{T0-P$_{bs}$}   &I think fat shaming is a form of bullying. \\
    \textbf{T0-P$_{gs}$}  &\textcolor{ForestGreen}{I think fatties are a problem.}\\
    \hline
    \textbf{GPS}        &\textcolor{ForestGreen}{this is just true: they are just talking like others do.} \\
    \textbf{GPT2-ft}  &\textit{If you think that shaming fat people can improve their lives you have a very limited and limited view of the world.} \\
    \textbf{DialoGPT-ft} & I'm not sure you know what'shaming' means.\\
    \hline
    \textbf{ReZG}   &I think the shaming of fat people is not only wrong, but it is also a form of violence, and \textcolor{violet}{it is a form in which the victim is not only dehumanized, but is also made to feel that they are not worthy of love}.\\
    \hline
  \end{tabular}
\label{tab_exp:case}
\end{table*}

\section{Ethical considerations}
This paper implements generating CNs with retrieved counter-knowledge in \stmdocstextcolor{black}{a zero-shot} paradigm, which \stmdocstextcolor{black}{improves} the quality of CNs and greatly reduces the manual workload. Although it is a promising approach to combat online HS without infringing on freedom of speech, the automatic generation of CNs inevitably faces offensive or disturbing content, which brings some ethical issues. Thus, we address the ethical considerations and remedy in terms of datasets, annotation, and generation tasks as follows:

\textbf{Annotation \stmdocstextcolor{black}{g}uidelines.} During the evaluation process, annotators will inevitably be exposed to offensive or disturbing content, which may affect their psychology. Thus, before starting the evaluation task, the annotators were trained and warned about the risk of being exposed to hate content. \stmdocstextcolor{black}{Additionally} they could withdraw from the task at any time if they were uncomfortable with the content. All annotators were adult volunteers, fully aware of the goals of the study.

\textbf{Dataset.} Since the datasets we chose were written and reviewed by experts, rather than personal conversation data collocated online, it avoids the threat to individual rights or personal privacy. In addition, the \stmdocstextcolor{black}{counter}-knowledge repository we used for retrieval is a forum with strict ethical regulations and prohibits rude or hostile comments, avoiding modeling inappropriate content like toxic content.

\textbf{Generation \stmdocstextcolor{black}{t}ask.} Although our method has substantially reduced the toxicity of CNs compared to existing methods, even the best models may still generate CNs containing toxicity or bias. To alleviate this problem, we suggest using post-filtering methods like toxicity detection and bias detection to filter out unqualified CNs, or measuring/promoting fairness in language models.

\section{Conclusion}
This \stmdocstextcolor{black}{study} focuses on \stmdocstextcolor{black}{the generation of} CNs with \stmdocstextcolor{black}{high} specificity in the absence of in-target annotated CN corpora. It presents ReZG, a novel retrieval-augmented zero-shot paradigm for CN generation. ReZG combines information retrieval and constrained decoding to obtain and incorporate counter-knowledge from external supporting materials \stmdocstextcolor{black}{in a flexible manner}. The introduction of differentiable constraints during the decoding phase enables more control over the generation process, allowing models to build complex relationships between \stmdocstextcolor{black}{the} retrieved counter-knowledge and target CNs.
The experimental results demonstrate that our method outperforms other state-of-the-art CN generation approaches with a stronger generalization capability, when generating out-target CNs.
It improves the countering \stmdocstextcolor{black}{success rate by 4.5\%+ compared to GPT2 and T0}. Although ReZG achieves outstanding performance, some limitations exist. First, while \stmdocstextcolor{black}{the time cost of} ReZG is considerably lower than human labor, gradient-based iterative decoding may be slower than single-recursive sample generation. Second, similar to many NLG tasks such as dialog generation, it relies on historical data and performs better on popular topics.
In future work, we plan to \stmdocstextcolor{black}{develop} the style-controlled CN generation methods \stmdocstextcolor{black}{further} and \stmdocstextcolor{black}{explore the means of enhancing the} model performance on less popular topics.

\section{Acknowledgements}
This work was supported by the National Natural Science Foundation of China under Grant Nos. 62202320 and 62302322, the Natural Science Foundation of Sichuan Province (No. 2024NSFSC1449), the Fundamental Research Funds for the Central Universities (Nos. 2022SCU12116, SCU2024D012), the Sichuan University Postdoctoral Interdisciplinary Innovation Startup Fund Project
(No. 10822041 A2076), and the Science and Engineering Connotation Development Project of Sichuan University
(No. 2020SCUNG129).

\appendix

\section{Stance embedding}
\label{sec:stance}
We fine-tune the pre-trained model \textsl{sup-simcse-bert-base-uncased} \footnotemark\footnotetext{\url{https://huggingface.co/princeton-nlp/sup-simcse-bert-base-uncased/}} ~\citep{gao2021simcse} with contrastive loss and stance detection datasets to obtain the embedding of stance in the vector space. \textsl{sup-simcse-bert-base-uncased} is an outstanding sentence embedding model which greatly improved the sentence embedding of BERT \citep{kenton2019bert} on semantic textual similarity tasks. We denote this fine-tuned model as \textsl{STA}.

Specifically, for each statement $s_{ i}$ in Vast ~\citep{allaway2020zero} and SemEval2016 Task-6 (TwitterStance) ~\citep{mohammad2016semeval} benchmark datasets, those that satisfy the following two conditions are constructed as positive pairs $\left(s_{i},s_{i}^{+}\right)$:
\begin{enumerate}
\item The targets of two statements are the same.
\item The polarities of two statements towards the target are the same.
\end{enumerate}
Statements with different stances for the same target are used as hard negative, $s_{i}^{-}$.
The contrastive loss function for the $\left(s_{i},s_{i}^{+},s_{i}^{-}\right)$ pairs is formulated as follows:
\begin{equation}
\begin{aligned}
\mathcal{L}_{i}= -\log \frac{e^{\operatorname{\textsl{STA}}\left(\mathbf{z}_{i}, \mathbf{z}_{i}^{+}\right) / \tau}}{\sum_{j=1}^{N}\left(e^{\operatorname{\textsl{STA}}\left(\mathbf{z}_{i}, \mathbf{z}_{j}^{+}\right) / \tau}+e^{\operatorname{\textsl{STA}}\left(\mathbf{z}_{i}, \mathbf{z}_{j}^{-}\right) / \tau}\right)},
\end{aligned}
\end{equation}
\begin{equation}
\begin{aligned}
\operatorname{\textsl{STA}}\left(\mathbf{z}_{i}, \mathbf{z}_{j}\right)=\frac{\mathbf{z}_{i}^{\top} \mathbf{z}_{j}}{\left\|\mathbf{z}_{i}\right\| \cdot\left\|\mathbf{z}_{j}\right\|}.
\label{eq2}
\end{aligned}
\end{equation}
$\textsl{STA}\left(\cdot,\cdot\right)$ represents the stance similarity between two statements, calculated by the cosine distance of two embedding vectors. $\tau$ is the temperature hyperparameter and $N$ is the size of the mini-batch. $\left(\mathbf{z}_{i},\mathbf{z}_{i}^{-}, \mathbf{z}_{i}^{+}\right)$ are the stance embedding vectors of $\left(s_{i},s_{i}^{+},s_{i}^{+}\right)$ respectively, which is calculated as follows:
\begin{equation}
\begin{array}{r}
\mathbf{z}_{i}=\text{MLP}\left(\text{BERT} \left(\texttt{[CLS]} \texttt{target} \texttt{[SEP]} s_{i} \texttt{[SEP]} \right)\right).
\end{array}
\end{equation}

Considering that a statement may have multiple targets and stances on every target could be different, every statement is formatted as $\texttt{[CLS]} \texttt{target} \texttt{[SEP]} s_{i} \texttt{[SEP]}$. $\texttt{[CLS]}$ is the pooling token, which is used to obtain the stance embedding for the entire statement. $\texttt{[SEP]}$ token is used to separating targets and statements.
\section{\textcolor{black}{Human Evaluation Examples}}
\label{sec:humam_example}
\textcolor{black}{Illustrations of the human evaluation results are shown in Tables \ref{tab:human_example1}, \ref{tab:human_example2} and \ref{tab:human-example3}.}

\begin{table*}[pos=htbp]
\centering
\caption{\textcolor{black}{Example 1 of the human evaluation results. The values in the array list represent the ratings provided by the three annotators, specifically: [Rating from Annotator 1, Rating from Annotator 2, Rating from Annotator 3]. The last row displays the Fleiss' Kappa values $k$ of the manual evaluation results under each metric, reflecting the consistency among the evaluation results.}}
\begin{tabular}{c|p{7cm}|c|c|c|c}
\hline
\rowcolor[HTML]{EFEFEF} 
\multicolumn{1}{c|}{\cellcolor[HTML]{EFEFEF}\textbf{Method}} & \multicolumn{1}{c|}{\cellcolor[HTML]{EFEFEF}\textbf{Content}}                                                                                                                                                                                    & \multicolumn{4}{c}{\cellcolor[HTML]{EFEFEF}\textbf{Evaluation Metrics}}                                                                                                              \\ \hline
\multicolumn{1}{c|}{HS}                                      & Children with Down Syndrome are a burden to society: they should be isolated, so that at least they do not hamper normal children's education.                                                                                                   & \multicolumn{1}{c|}{\textbf{LQ}}                   & \multicolumn{1}{c|}{\textbf{SPE}}                  & \multicolumn{1}{c|}{\textbf{CON}}                  & \textbf{BEST}         \\ \hline
\multicolumn{1}{c|}{GPS}                                     & Despite the negative stereotypes, many children with Down Syndrome enjoy regular playtime and interacting with other children.                                                                                                                   & \multicolumn{1}{c|}{[5, 5, 5]}         & \multicolumn{1}{c|}{[2, 3, 3]}         & \multicolumn{1}{c|}{[0, 0, 0]}         & [0, 0, 0] \\
\multicolumn{1}{c|}{GPT2-ft}                                 & This is a very sad and worrying statement: that children with Down syndrome are a burden to society, and that they should be isolated from others. We should be working to ensure that everyone is able to live and study in a safe environment. & \multicolumn{1}{c|}{[4, 4, 3]}         & \multicolumn{1}{c|}{[4, 4, 3]}         & \multicolumn{1}{c|}{[1, 1, 1]}         & [0, 0, 0] \\
\multicolumn{1}{c|}{DialoGPT-ft}                             & Children with Down Syndrome are often bullied, bullied, bullied on their behalf and they should not be isolated.                                                                                                                                 & \multicolumn{1}{c|}{[3, 4, 3]}         & \multicolumn{1}{c|}{[2, 3, 2]}         & \multicolumn{1}{c|}{[0, 0, 0]}         & [0, 0, 0] \\ \hline
\multicolumn{1}{c|}{GPT2-P$_{topk}$}                         & I do not believe that the child should be isolated.                                                                                                                                                                                              & \multicolumn{1}{c|}{[5, 5, 5]}         & \multicolumn{1}{c|}{[2, 2, 1]}         & \multicolumn{1}{c|}{[1, 1, 0]}         & [0, 0, 0] \\
\multicolumn{1}{c|}{GPT2-P$_{topp}$}                         & I believe that we should provide every child with the best education possible, regardless of their disability.                                                                                                                                   & \multicolumn{1}{c|}{[5, 5, 5]}         & \multicolumn{1}{c|}{[2, 3, 2]}         & \multicolumn{1}{c|}{[1, 1, 1]}         & [0, 0, 0] \\
\multicolumn{1}{c|}{GPT2-P$_{bs}$}                           & I believe that children with Down syndrome should be given the same opportunities as other children.                                                                                                                                             & \multicolumn{1}{c|}{[5, 5, 5]}         & \multicolumn{1}{c|}{[2, 4, 4]}         & \multicolumn{1}{c|}{[1, 1, 1]}         & [0, 0, 0] \\
\multicolumn{1}{c|}{GPT2-P$_{gs}$}                           & I believe that the best way to help children with Down syndrome is to give them the best education possible.                                                                                                                                     & \multicolumn{1}{c|}{[5, 5, 5]}         & \multicolumn{1}{c|}{[3, 3, 3]}         & \multicolumn{1}{c|}{[0, 1, 0]}         & [0, 0, 0] \\
\multicolumn{1}{c|}{T0-P$_{topk}$}                           & I think that people with Down Syndrome should be included as much as possible. Down Syndrome is a genetic condition, but not a disease.                                                                                                          & \multicolumn{1}{c|}{[5, 4, 4]}         & \multicolumn{1}{c|}{[3, 3, 3]}         & \multicolumn{1}{c|}{[1, 1, 1]}         & [0, 0, 0] \\
\multicolumn{1}{c|}{T0-P$_{topp}$}                           & Down Syndrome children are a joy, and they deserve to be included in society.                                                                                                                                                                    & \multicolumn{1}{c|}{[5, 5, 5]}         & \multicolumn{1}{c|}{[3, 3, 3]}         & \multicolumn{1}{c|}{[1, 1, 1]}         & [0, 0, 0] \\
\multicolumn{1}{c|}{T0-P$_{bs}$}                             & Down syndrome is a genetic disorder that affects one in every 6,500 live births.                                                                                                                                                                 & \multicolumn{1}{c|}{[5, 5, 5]}         & \multicolumn{1}{c|}{[0, 0, 0]}         & \multicolumn{1}{c|}{[0, 0, 0]}         & [0, 0, 0] \\
\multicolumn{1}{c|}{T0-P$_{gs}$}                             & I disagree, because they are not a burden to society. They are a blessing.                                                                                                                                                                       & \multicolumn{1}{c|}{[4, 5, 4]}         & \multicolumn{1}{c|}{[3, 4, 4]}         & \multicolumn{1}{c|}{[1, 1, 1]}         & [0, 0, 0] \\ \hline
\multicolumn{1}{c|}{ReZG}                                    & I believe that the majority of people with Down Syndrome are not a burden to society. They bring unique perspectives and value and should be treated as equals.                                                                                  & \multicolumn{1}{c|}{[5, 5, 5]}         & \multicolumn{1}{c|}{[4, 4, 4]}         & \multicolumn{1}{c|}{[1, 1, 1]}         & [1, 1, 1] \\ \hline
\rowcolor[HTML]{EFEFEF} 
\multicolumn{2}{c|}{\cellcolor[HTML]{EFEFEF}Fleiss' Kappa: $k$}    & \multicolumn{1}{c|}{\cellcolor[HTML]{EFEFEF}0.607} & \multicolumn{1}{c|}{\cellcolor[HTML]{EFEFEF}0.457} & \multicolumn{1}{c|}{\cellcolor[HTML]{EFEFEF}0.745} & 1.000                 \\ \hline
\end{tabular}
\label{tab:human_example1}
\end{table*}

\begin{table*}[pos=htbp]
\centering
\caption{\textcolor{black}{Example 2 of the human evaluation results. The values in the array list represent the ratings provided by the three annotators, specifically: [Rating from Annotator 1, Rating from Annotator 2, Rating from Annotator 3]. The last row displays the Fleiss' Kappa values $k$ of the manual evaluation results under each metric, reflecting the consistency among the evaluation results.}}
\begin{tabular}{cp{7cm}|cccc}
\hline
\rowcolor[HTML]{EFEFEF} 
\multicolumn{1}{c|}{\cellcolor[HTML]{EFEFEF}\textbf{Method}} & \multicolumn{1}{c|}{\cellcolor[HTML]{EFEFEF}\textbf{Content}}                                                                                                               & \multicolumn{4}{c}{\cellcolor[HTML]{EFEFEF}\textbf{Evaluation Metrics}}                                                                      \\ \hline
\multicolumn{1}{c|}{HS}                                      & Shaming Fatties is good to make them   taking healthier habits.                                                                                                             & \multicolumn{1}{c|}{\textbf{LQ}} & \multicolumn{1}{c|}{\textbf{SPE}} & \multicolumn{1}{c|}{\textbf{CON}} & \multicolumn{1}{c}{\textbf{BEST}} \\ \hline
\multicolumn{1}{c|}{GPS}                                     & these   words are taking trouble against based on your sexuality.                                                                                                           & \multicolumn{1}{l|}{{[}3, 3, 4{]}} & \multicolumn{1}{l|}{{[}1, 1, 1{]}}  & \multicolumn{1}{l|}{{[}0, 0, 0{]}}  & {[}0, 0, 0{]}                       \\ \hline
\multicolumn{1}{c|}{GPT2-ft}                                 & I   don't see how shaming people for their physical characteristics will help   anyone to make healthier choices.                                                           & \multicolumn{1}{l|}{{[}5, 5, 5{]}} & \multicolumn{1}{l|}{{[}3, 3, 3{]}}  & \multicolumn{1}{l|}{{[}1, 1, 1{]}}  & {[}0, 0, 0{]}                       \\ \hline
\multicolumn{1}{c|}{DialoGPT-ft}                             & Fatties have a lot more in common with overweight   people than they do with healthy ones.                                                                                  & \multicolumn{1}{l|}{{[}5, 5, 5{]}} & \multicolumn{1}{l|}{{[}0, 0, 0{]}}  & \multicolumn{1}{l|}{{[}0, 0, 0{]}}  & {[}0, 0, 0{]}                       \\ \hline
\multicolumn{1}{c|}{GPT2-P$_{topk}$}                               & I   don't think you should shame them into eating healthier,  I think you should   make them aware of their body image issues and how to fix them if                         & \multicolumn{1}{l|}{{[}3, 4, 3{]}} & \multicolumn{1}{l|}{{[}4, 4, 3{]}}  & \multicolumn{1}{l|}{{[}1, 1, 0{]}}  & {[}0, 0, 0{]}                       \\ \hline
\multicolumn{1}{c|}{GPT2-P$_{topp}$}                               & I   don't think shaming is a good way to get people to change their behavior.   It's like shaming a fat person to lose weight.                                              & \multicolumn{1}{l|}{{[}5, 5, 5{]}} & \multicolumn{1}{l|}{{[}3, 3, 3{]}}  & \multicolumn{1}{l|}{{[}1, 1, 1{]}}  & {[}0, 0, 0{]}                       \\ \hline
\multicolumn{1}{c|}{GPT2-P$_{bs}$}                                 & I   think shaming them is not the right way to go about it. I think shaming them   is not the right way to go about it.                                                     & \multicolumn{1}{l|}{{[}3, 3, 3{]}} & \multicolumn{1}{l|}{{[}3, 3, 3{]}}  & \multicolumn{1}{l|}{{[}0, 0, 0{]}}  & {[}0, 0, 0{]}                       \\ \hline
\multicolumn{1}{c|}{GPT2-P$_{gs}$}                                 & I   think shaming is a very bad thing. I think shaming is a very bad thing.                                                                                                 & \multicolumn{1}{l|}{{[}3, 3, 3{]}} & \multicolumn{1}{l|}{{[}3, 3, 3{]}}  & \multicolumn{1}{l|}{{[}0, 0, 0{]}}  & {[}0, 0, 0{]}                       \\ \hline
\multicolumn{1}{c|}{T0-P$_{topk}$}                                 & It   makes people feel bad about themselves..                                                                                                                               & \multicolumn{1}{l|}{{[}4, 4, 5{]}} & \multicolumn{1}{l|}{{[}3, 4, 3{]}}  & \multicolumn{1}{l|}{{[}1, 1, 1{]}}  & {[}0, 0, 0{]}                       \\ \hline
\multicolumn{1}{c|}{T0-P$_{topp}$}                                 & Shaming   Fatties is bad for their self-esteem.                                                                                                                             & \multicolumn{1}{l|}{{[}5, 5, 5{]}} & \multicolumn{1}{l|}{{[}4, 4, 3{]}}  & \multicolumn{1}{l|}{{[}1, 1, 1{]}}  & {[}0, 0, 0{]}                       \\ \hline
\multicolumn{1}{c|}{T0-P$_{bs}$}                                   & In   my opinion, shaming fatties is not good to make them take healthier habits.                                                                                            & \multicolumn{1}{l|}{{[}5, 5, 5{]}} & \multicolumn{1}{l|}{{[}3, 3, 3{]}}  & \multicolumn{1}{l|}{{[}1, 1, 1{]}}  & {[}0, 0, 0{]}                       \\ \hline
\multicolumn{1}{c|}{T0-P$_{gs}$}                                   & I   think fat people should be praised for their efforts to lose weight.                                                                                                    & \multicolumn{1}{l|}{{[}5, 5, 5{]}} & \multicolumn{1}{l|}{{[}3, 2, 2{]}}  & \multicolumn{1}{l|}{{[}1, 1, 1{]}}  & {[}0, 0, 0{]}                       \\ \hline
\multicolumn{1}{c|}{ReZG}                                    & Those who are shamed may internalize the shame and   believe they are a bad person,  maybe encouraging them to have healthy habits but   not humiliating is a better choice. & \multicolumn{1}{l|}{{[}4, 5, 5{]}} & \multicolumn{1}{l|}{{[}4, 4, 5{]}}  & \multicolumn{1}{l|}{{[}1, 1, 1{]}}  & {[}1, 1, 1{]}                       \\ \hline
\rowcolor[HTML]{EFEFEF} 
\multicolumn{2}{c|}{\cellcolor[HTML]{EFEFEF}Fleiss’ Kappa: $k$}                                                                                                                                                                            & \multicolumn{1}{C|}{\cellcolor[HTML]{EFEFEF}0.605} & \multicolumn{1}{C|}{\cellcolor[HTML]{EFEFEF}0.563} & \multicolumn{1}{C|}{\cellcolor[HTML]{EFEFEF}0.880} & 1.000                               \\ \hline
\end{tabular}
\label{tab:human_example2}
\end{table*}

\begin{table*}[pos=htbp]
\centering
\caption{\textcolor{black}{Example 3 of the human evaluation results. The values in the array list represent the ratings provided by the three annotators, specifically: [Rating from Annotator 1, Rating from Annotator 2, Rating from Annotator 3]. The last row displays the Fleiss' Kappa values $k$ of the manual evaluation results under each metric, reflecting the consistency among the evaluation results.}}
\begin{tabular}{cp{7cm}|cccc}
\hline
\rowcolor[HTML]{EFEFEF} 
\multicolumn{1}{c|}{\cellcolor[HTML]{EFEFEF}\textbf{Method}} & \multicolumn{1}{c|}{\cellcolor[HTML]{EFEFEF}\textbf{Content}}                                                                                                                                                          & \multicolumn{4}{c}{\cellcolor[HTML]{EFEFEF}\textbf{Evaluation Metrics}}                                                                                                      \\ \hline
\multicolumn{1}{c|}{HS}                                      & Shaming Fatties is just for their   good.                                                                                                                                                                              & \multicolumn{1}{c|}{\textbf{LQ}}                   & \multicolumn{1}{c|}{\textbf{SPE}}                  & \multicolumn{1}{c|}{\textbf{CON}}                  & \textbf{BEST} \\ \hline
\multicolumn{1}{c|}{GPS}                                     & this   is just true: they are just talking like others do                                                                                                                                                              & \multicolumn{1}{c|}{{[}4, 4, 4{]}}                   & \multicolumn{1}{c|}{{[}1, 1, 0{]}}                   & \multicolumn{1}{c|}{{[}0, 0, 0{]}}                   & {[}0, 0, 0{]}   \\ \hline
\multicolumn{1}{c|}{GPT2-ft}                                 & If   you think that shaming fat people can improve their lives you have a very limited and limited view of the world.                                                                                                  & \multicolumn{1}{c|}{{[}3, 4, 3{]}}                   & \multicolumn{1}{c|}{{[}3, 3, 2{]}}                   & \multicolumn{1}{c|}{{[}1, 1, 1{]}}                   & {[}0, 0, 0{]}   \\ \hline
\multicolumn{1}{c|}{DialoGPT-ft}                                & I’m   not sure you know what’ shaming’ means.                                                                                                                                                                          & \multicolumn{1}{c|}{{[}4, 4, 4{]}}                   & \multicolumn{1}{c|}{{[}2, 2, 2{]}}                   & \multicolumn{1}{c|}{{[}1, 0, 0{]}}                   & {[}0, 0, 0{]}   \\ \hline
\multicolumn{1}{c|}{GPT2-P$_{topk}$}                               & I   don’t want to shame the   obese,  I don’t want to shame the   fat people.                                                                                                                                           & \multicolumn{1}{c|}{{[}3, 3, 2{]}}                   & \multicolumn{1}{c|}{{[}2, 2, 3{]}}                   & \multicolumn{1}{c|}{{[}1, 0, 0{]}}                   & {[}0, 0, 0{]}   \\ \hline
\multicolumn{1}{c|}{GPT2-P$_{topp}$}                               & I   don’ t like shaming anyone. It’ s not my place to make that decision for   anyone                                                                                                                                  & \multicolumn{1}{c|}{{[}5, 5, 5{]}}                   & \multicolumn{1}{c|}{{[}3, 3, 3{]}}                   & \multicolumn{1}{c|}{{[}1, 1, 1{]}}                   & {[}0, 0, 0{]}   \\ \hline
\multicolumn{1}{c|}{GPT2-P$_{bs}$}                                 & I   think shaming people for being fat is just for their good.                                                                                                                                                         & \multicolumn{1}{c|}{{[}5, 5, 5{]}}                   & \multicolumn{1}{c|}{{[}0, 0, 0{]}}                   & \multicolumn{1}{c|}{{[}0, 0, 0{]}}                   & {[}0, 0, 0{]}   \\ \hline
\multicolumn{1}{c|}{GPT2-P$_{gs}$}                                 & I   think shaming is a very effective way to get people to change their behavior.                                                                                                                                      & \multicolumn{1}{c|}{{[}5, 5, 5{]}}                   & \multicolumn{1}{c|}{{[}0, 0, 0{]}}                   & \multicolumn{1}{c|}{{[}0, 0, 0{]}}                   & {[}0, 0, 0{]}   \\ \hline
\multicolumn{1}{c|}{T0-P$_{topk}$}                                 & I   don't believe fat people should be shaming.                                                                                                                                                                        & \multicolumn{1}{c|}{{[}4, 4, 3{]}}                   & \multicolumn{1}{c|}{{[}2, 2, 2{]}}                   & \multicolumn{1}{c|}{{[}1, 1, 1{]}}                   & {[}0, 0, 0{]}   \\ \hline
\multicolumn{1}{c|}{T0-P$_{topp}$}                                 & It's   degrading to the overweight and obese.                                                                                                                                                                          & \multicolumn{1}{c|}{{[}5, 5, 4{]}}                   & \multicolumn{1}{c|}{{[}1, 1, 2{]}}                   & \multicolumn{1}{c|}{{[}0, 0, 0{]}}                   & {[}0, 0, 0{]}   \\ \hline
\multicolumn{1}{c|}{T0-P$_{bs}$}                                   & I   think fat shaming is a form of bullying.                                                                                                                                                                           & \multicolumn{1}{c|}{{[}5, 5, 5{]}}                   & \multicolumn{1}{c|}{{[}3, 3, 3{]}}                   & \multicolumn{1}{c|}{{[}1, 1, 1{]}}                   & {[}0, 0, 0{]}   \\ \hline
\multicolumn{1}{c|}{T0-P$_{gs}$}                                   & I   think fatties are a problem.                                                                                                                                                                                       & \multicolumn{1}{c|}{{[}5, 5, 5{]}}                   & \multicolumn{1}{c|}{{[}0, 0, 0{]}}                   & \multicolumn{1}{c|}{{[}0, 0, 0{]}}                   & {[}0, 0, 0{]}   \\ \hline
\multicolumn{1}{c|}{ReZG}                                    & I   think the shaming of fat people is not only wrong,  but it is also a form of   violence,  and it is a form in which the victim is not only dehumanized,  but   is also made to feel that they are not worthy of love. & \multicolumn{1}{c|}{{[}4, 5, 5{]}}                   & \multicolumn{1}{c|}{{[}4, 4, 5{]}}                   & \multicolumn{1}{c|}{{[}1, 1, 1{]}}    & {[}1, 1 1{]}   \\ \hline
\rowcolor[HTML]{EFEFEF} 
\multicolumn{2}{c|}{\cellcolor[HTML]{EFEFEF}Fleiss’ Kappa: $k$}                       & \multicolumn{1}{c|}{\cellcolor[HTML]{EFEFEF}0.543} & \multicolumn{1}{c|}{\cellcolor[HTML]{EFEFEF}0.641} & \multicolumn{1}{c|}{\cellcolor[HTML]{EFEFEF}0.777} & 1.000         \\ \hline
\end{tabular}
\label{tab:human-example3}
\end{table*}

\section{Qualification questionnaire}
\label{sec:qq}
\begin{figure*}[h!]
  \centering
  \includegraphics[width=1.\textwidth]{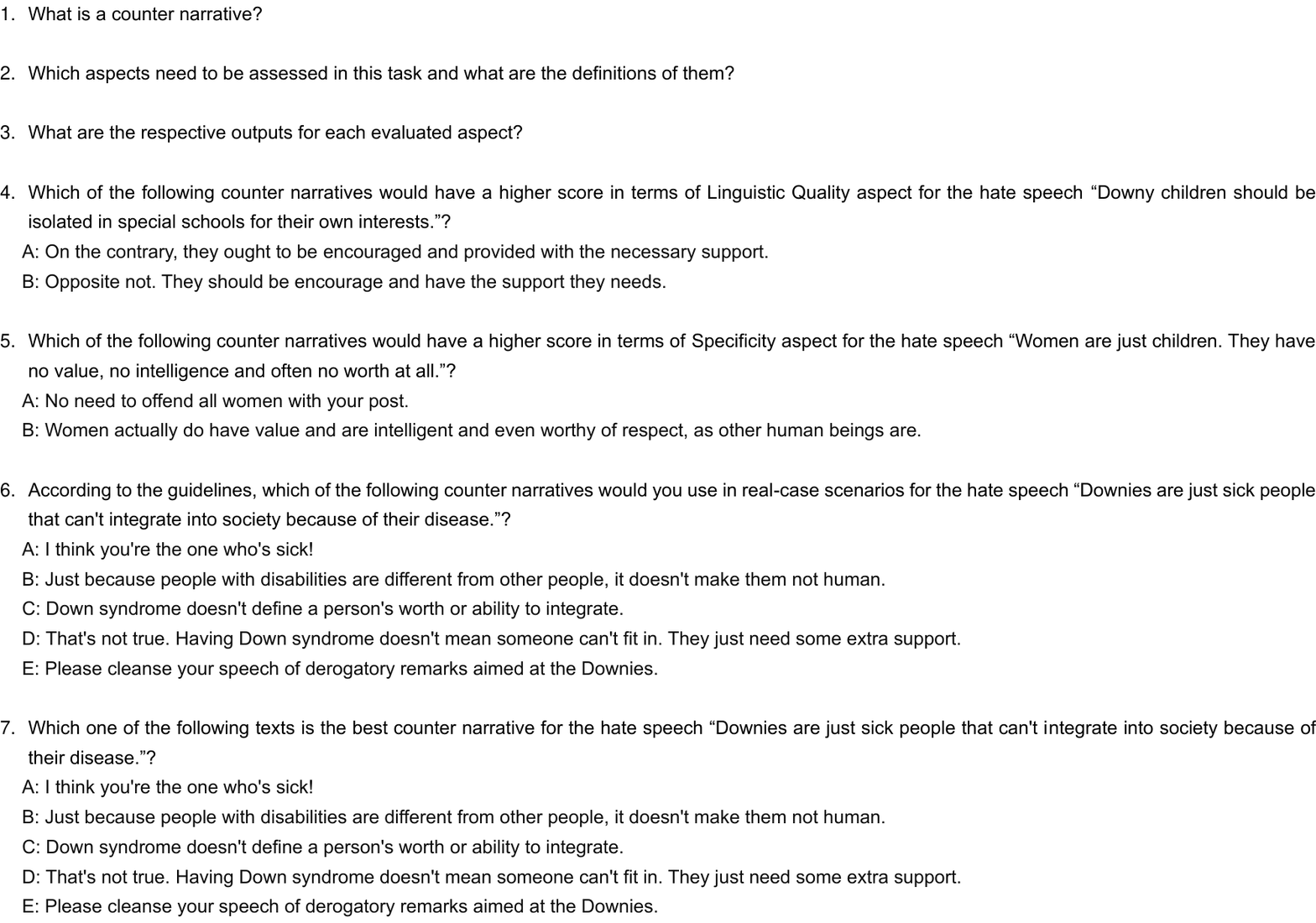}
\caption{ The qualification questionnaire for annotators.}
  \label{exp:qq}
\end{figure*}

See Fig. \ref{exp:qq}.

\printcredits
\section{\textcolor{black}{Prompts used to generate CNs with the retrieved information}}
\label{sec:eng_prompt}
\textcolor{black}{The prompts used to generate CNs with the retrieved information in Section \ref{sec:eng_ex}
are shown in Table \ref{tab_exp:eng_prompt}.}

\begin{table*}[b!]
\centering
\caption{\textcolor{black}{The prompts used in the ``\textit{EnG vs. other generators}'' experiment in Section \ref{sec:eng_ex}. These prompts are designed to guide LMs in generating CNs using the retrieved external information.}}
 \begin{tabular}{c|p{13cm}}
 \hline
 \textbf{Prompt}&\textbf{Content} \\
    \hline
   SimPrompt &Here is the hate speech: <Hate Speech>. 
   
     Considering that <Retrieved Information>, I disagree with the above hate speech because: \\
     \hline
    EnGPrompt &Your task is to generate a counter narrative to the given \#Hate Speech\# based on the \#Retrieved Information\#. The counter narrative must meet the following requirements:
\newline
1. Retain counter-knowledge: Retain as much of the retrieved information as possible that can be used to counter the hate speech, ensuring that this knowledge is aligned with the core logic of the rebuttal.
\newline
2. Maintain fluency: The counter narrative should be natural and coherent, avoiding awkward or unclear expressions.
\newline
3. Counter the given \#Hate Speech\#.
\newline \newline
Below are the given \#Hate Speech\# and the \#Retrieved Information\#. Please return the generated counter narrative text after \#Counter Narrative\#, and do not return any other content.\newline
\#Hate Speech\#: <Hate Speech>. \newline
\#Retrieved Information\#: <Retrieved Information>.\newline
\#Counter Narrative\#: \\
     \hline 
 KnoPrompt&Topic: Online shopping. User says:  I love using Amazon, have you tried it? We know that: Online shopping is the process of purchasing goods or services online from a website or other online  store. System replies: Yes I have. I love using Amazon. I know that Online shopping is the process of purchasing goods or services from a website or other online service provider.
\newline \newline
Topic: inhaling helium. User says:  I did it from the balloon lol. I have trouble inhaling it for some reason. Is there any long-term risks with helium inhalation? We know that: The long-term risks for inhaling helium include: shortness of breath, chest tightness, and coughing. System replies: Yeah , I know that long term risks for inhaling helium includes shortness of breath, chest tightness, and coughing.
\newline \newline
Topic: Kyoto. User says: Good morning, this is my first day visiting japan. I've since kyoto in many animes and would love to see it in person. We know that: Kyoto is considered the cultural capital of Japan. System replies: Great ! I remember Kyoto is considered the cultural capital of Japan.
\newline
Topic: <The target of hate speech>. User says: <Hate Speech>. We know that: <Retrieved Information>. System replies:\\
\hline
  \end{tabular}
\label{tab_exp:eng_prompt}
\end{table*}
\bibliographystyle{model1-num-names}
\bibliography{Reference_clean}
\end{document}